%% file: p.tex
\documentclass[sigconf, screen]{acmart}

\newcommand{\paperTitle}{Zeus: Efficiently Localizing Actions in Videos using
Reinforcement Learning}

\usepackage{anyfontsize}
\usepackage[hyphens]{url}

\definecolor{linkcolor}{HTML}{647382}
\definecolor{citecolor}{HTML}{647382} %
\definecolor{urlcolor}{rgb}{0.4,0.2,0.2}
\definecolor{sqlcolor}{HTML}{965d67}
\definecolor{smtcolor}{HTML}{5d968c}
\definecolor{webblue}{rgb}{0,0,.7}
\definecolor{webred}{rgb}{0.8,0,0}
\definecolor{webgreen}{rgb}{0,.5,0}
\definecolor{webbrown}{rgb}{.6,0,0}

\usepackage{amsmath,amsopn,amssymb}
\usepackage[cal=boondoxo]{mathalfa}
\usepackage{subcaption}
\usepackage{endnotes,microtype,xspace,graphicx,fancyvrb,multirow}
\usepackage{supertabular,booktabs}
\usepackage{array,underscore, relsize}
\usepackage{fancyhdr}
\usepackage{enumitem}
\usepackage{balance}
\usepackage{booktabs}
\usepackage{pifont}
\usepackage{listings}
\usepackage{multirow}
\usepackage[scaled]{beramono}
\usepackage{tabularx}
\usepackage{multirow}

\makeatletter
\newcommand\BeraMonottfamily{%
  \def\fvm@Scale{0.85}
  \fontfamily{fvm}\selectfont
}
\makeatother

\definecolor{mymauve}{rgb}{0.58,0,0.82}

\lstdefinestyle{SQLStyle}{
  language=SQL,
  basicstyle={\small\ttfamily},
  breaklines=true,
  frame=none,
  numbers=none,
  keepspaces=true,
  captionpos=b,
  stringstyle=\color{mymauve},
  keywordstyle=\color{blue},
  commentstyle=\color{dkgreen},
}

\lstdefinestyle{ScriStyle}{
language=SQL,
basicstyle=\BeraMonottfamily\footnotesize, 
keywordstyle=\color{smtcolor}\bfseries,
morekeywords={and, or, not},
aboveskip = 0.05in,
belowskip = 0.05in,
literate = {-}{-}1, 
}

\usepackage{aliascnt}

\usepackage[noend, ruled,linesnumbered]{algorithm2e}
\usepackage{setspace}
\usepackage{enumitem}

\SetAlFnt{\small}
\SetAlCapFnt{\small}
\SetAlCapNameFnt{\small}

\usepackage{semantic}

\usepackage{stmaryrd}
\usepackage{ltablex}
\usepackage{mathtools}
\usepackage{adjustbox}
\usepackage{fixltx2e}
\usepackage{bibentry}
\usepackage[group-separator={,}]{siunitx}
\newcommand{\sys}{\mbox{\textsc{Zeus}}\xspace}

\newcommand{\samplingrate}{\mbox{\textsc{Sampling Rate}}\xspace}
\newcommand{\segmentlength}{\mbox{\textsc{Segment Length}}\xspace}
\newcommand{\resolution}{\mbox{\textsc{Resolution}}\xspace}
\newcommand{\configuration}{\mbox{\textsc{Configuration}}\xspace}

\newcommand{\crossleft}{\mbox{\textsc{CrossLeft}}\xspace}
\newcommand{\crossright}{\mbox{\textsc{CrossRight}}\xspace}

\newcommand{\leftturn}{\mbox{\textsc{LeftTurn}}\xspace}

\newcommand{\polevault}{\mbox{\textsc{PoleVault}}\xspace}
\newcommand{\cleanandjerk}{\mbox{\textsc{CleanAndJerk}}\xspace}
\newcommand{\ironingclothes}{\mbox{\textsc{IroningClothes}}\xspace}
\newcommand{\tennisserve}{\mbox{\textsc{TennisServe}}\xspace}

\newcommand{\framepp}{\mbox{\textsc{Frame-PP}}\xspace}
\newcommand{\segmentpp}{\mbox{\textsc{Segment-PP}}\xspace}
\newcommand{\windowpp}{\mbox{\textsc{Zeus-Sliding}}\xspace}
\newcommand{\heuristic}{\mbox{\textsc{Zeus-Heuristic}}\xspace}
\newcommand{\sysrl}{\mbox{\textsc{Zeus-RL}}\xspace}

\usepackage[capitalize,noabbrev,nameinlink]{cleveref}

\crefname{lstlisting}{listing}{listings}
\Crefname{lstlisting}{Listing}{Listings}

\input{cmds}



\usepackage{chngcntr}
\counterwithout{equation}{section}

\copyrightyear{2022}
\acmYear{2022}
\setcopyright{rightsretained}

\acmConference[SIGMOD '22]{Proceedings of the 2022 International Conference on Management of Data}{June 12--17, 2022}{Philadelphia, PA, USA.}
\acmBooktitle{Proceedings of the 2022 International Conference on Management of Data (SIGMOD '22), June 12--17, 2022, Philadelphia, PA, USA}
\acmDOI{10.1145/3514221.3526181}
\acmISBN{978-1-4503-9249-5/22/06.}

\usepackage{etoolbox}
\makeatletter
\patchcmd{\maketitle}{\@copyrightpermission}{
   \begin{minipage}{0.4\columnwidth}
     \href{http://creativecommons.org/licenses/by/4.0/}{\includegraphics[width=0.95\textwidth]{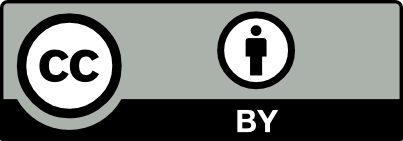}}
   \end{minipage}\hfill
   \begin{minipage}{0.6\columnwidth}
     \href{http://creativecommons.org/licenses/by/4.0/}{This work is licensed under a Creative Commons Attribution International 4.0 License.}
   \end{minipage}
 
   \vspace{5pt}
}{}{}

\makeatother



\title{\paperTitle} 

\begin{document}
\fancyhead{}
\author{Pramod Chunduri}
\affiliation{%
  \institution{Georgia Institute of Technology}
}
\email{pramodc@gatech.edu}

\author{Jaeho Bang}
\affiliation{%
  \institution{Georgia Institute of Technology}
}
\email{jaehobang@gatech.edu}

\author{Yao Lu}
\affiliation{%
  \institution{Microsoft Research}
}
\email{luyao@microsoft.com}

\author{Joy Arulraj}
\affiliation{%
  \institution{Georgia Institute of Technology}
}
\email{arulraj@gatech.edu}

\input{abstract}
\maketitle





\input{introduction}

\input{overview}

\input{execution}

\input{planning}

\input{implementation}
\input{evaluation}
\input{related-work}

\input{conclusion}



\balance
\bibliographystyle{ACM-Reference-Format}
\raggedright
\bibliography{p}

\end{document}

%% file: cmds.tex
\fvset{fontsize=\scriptsize,xleftmargin=8pt,numbers=left,numbersep=5pt}

\input{code/fmt}

\definecolor{dkgreen}{rgb}{0,0.6,0}

\def\Snospace~{\S{}}

\numberwithin{equation}{section}

\newif\ifdraft\drafttrue
\newif\ifnotes\notestrue
\ifdraft\else\notesfalse\fi

\newcommand{\rev}[1]{{#1}}

\input{glyphtounicode}
\newcolumntype{R}[1]{>{\raggedleft\let\newline\\\arraybackslash\hspace{0pt}}p{#1}}

\SetCommentSty{mycommfont}

\usepackage{booktabs,tikz}
\newcommand\tikzmark[1]{\tikz[remember picture] \node (#1) {};}

\newcommand{\PP}[1]{
\vspace{2px}
\noindent{\bf\textsc{#1}.}\xspace
}

\newcommand{\squishitemize}{
 \begin{list}{$\bullet$}
  { \setlength{\itemsep}{0pt}
     \setlength{\parsep}{3pt}
     \setlength{\topsep}{0pt}
     \setlength{\partopsep}{0pt}
     \setlength{\leftmargin}{1.95em}
     \setlength{\labelwidth}{1.5em}
     \setlength{\labelsep}{0.5em} } }

\newcounter{Lcount}
\newcommand{\squishlist}{
    \begin{list}{\arabic{Lcount}. }
   { \usecounter{Lcount}
        \setlength{\itemsep}{0pt}
        \setlength{\parsep}{3pt}
        \setlength{\topsep}{0pt}
        \setlength{\partopsep}{0pt}
        \setlength{\leftmargin}{2em}
        \setlength{\labelwidth}{1.5em}
        \setlength{\labelsep}{0.5em} } }

\newcommand{\squishend}{\end{list}}

\usepackage{xstring}

\newcommand{\eg}{\textit{e}.\textit{g}.,\xspace}
\newcommand{\ie}{\textit{i}.\textit{e}.,\xspace}

\newcommand{\abbrev}{\textit{abbrev}.,~}

\newcommand{\opt}{\textsc{query planner}\xspace}
\newcommand{\proc}{\textsc{query executor}\xspace}

\newcommand{\rl}{\textsc{RL}\xspace}
\newcommand{\udf}{\textsc{UDF}\xspace}
\newcommand{\apfg}{\textsc{APFG}\xspace}
\newcommand{\feature}{\textsc{ProxyFeature}\xspace}

\newcommand{\qa}{\texttt{Q1}\xspace}

\newcommand{\qf}{\texttt{Q6}\xspace}

\newcommand{\rthreed}{\texttt{\textsc{R3D}}\xspace}

\newcommand{\cmark}{\ding{51}}

\newcommand{\rpm}{\sbox0{$1$}\sbox2{$\scriptstyle\pm$}
  \raise\dimexpr(\ht0-\ht2)/2\relax\box2 }

\newcommand{\RomanNumeralCaps}[1]
    {\MakeUppercase{\romannumeral #1}}

%% file: abstract.tex
\begin{abstract}
Detection and localization of actions in videos is an important problem in
practice.
State-of-the-art video analytics systems are unable to efficiently and
effectively answer such action queries because
actions often involve a complex interaction between objects and are spread across
a sequence of frames;
detecting and localizing them requires computationally expensive
deep neural networks.
It is also important to consider the entire sequence of frames to
answer the query effectively.

In this paper, we present \sys, a video analytics system tailored for answering 
action queries.
We present a novel technique for efficiently answering these queries using deep
reinforcement learning.
\sys trains a reinforcement learning agent that learns to adaptively modify 
the input video segments that are subsequently sent to an action classification
network.
The agent alters the input segments along three dimensions - sampling rate, 
segment length, and resolution.
To meet the user-specified accuracy target, \sys{'s} query optimizer trains the
agent based on an accuracy-aware, aggregate reward function.
Evaluation on three diverse video datasets shows that \sys
outperforms state-of-the-art frame- and window-based filtering techniques
by up to 22.1$\times$ and 4.7$\times$, respectively. 
It also consistently meets the user-specified accuracy target across all
queries. \looseness=-1
\end{abstract}

%% file: introduction.tex
\section{Introduction}
\label{sec:introduction}

\begin{figure}
    \centering
    \begin{subfigure}[b]{\columnwidth}
        \centering
        \includegraphics[width=\linewidth]{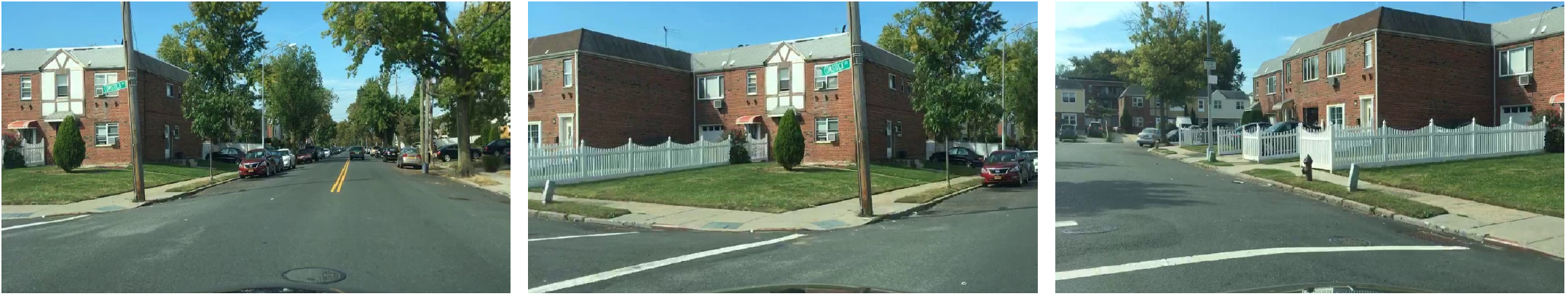}
        \label{fig:left-turn}
    \end{subfigure}
    \caption{ 
        Sequence of frames capturing a car taking a left turn.
        None of the individual frames are sufficient to
        independently determine and localize the action.
        State-of-the-art frame-level techniques tailored for object
        detection are unable to answer action queries.
    }
        \label{fig:left-right-turn}
\end{figure}

%
Recent advances in video database management systems
(VDBMSs) \cite{kang13blazeit,lu2018accelerating,bastani2020miris} have enabled automated analysis of videos at scale.
These systems have primarily focused on retrieving frames that contain an object
of interest.
We instead are interested in detection and localization of \textit{actions} in long,
untrimmed videos~\cite{shou2016temporal,
xia2020survey, chao2018rethinking}.
For example, as shown in~\cref{fig:left-right-turn}, an action refers to an event spread across a
sequence of frames. 
A traffic analyst might be interested in studying the patterns in
which vehicles move at a given intersection.
They might want to identify the density of pedestrian crossing in a particular 
direction.

\PP{Action Localization Task}
Action localization involves locating the start and end point of an action in a
video, and classifying those frames into
one of the available action classes (\eg `left turn of a car').
Processing \textit{action queries} requires the detection and localization
of all occurrences of a given action in the video.
The following query retrieves all the video \textit{segments}
(\ie a contiguous sequence of frames) that contain a left turn of a car 
with 80\% accuracy:
\begin{lstlisting}[style=SQLStyle,
                   morekeywords={PRODUCE, PROCESS},
                   label={lst:sql-query-a}]
--- Retrieve segments with a left turn action
SELECT segment_ids FROM UDF(video)
WHERE action_class = 'left-turn'
AND accuracy >= 80%
\end{lstlisting}

In this query, $segment\_id$ is a unique identifier for each segment in the
video with a start and end frame;
$accuracy$ is the user-specified target accuracy for the query.
The \udf is a user-defined function that returns predictions for each segment including
the action class (\eg turn) and the action boundary (list of segments).

\begin{figure}
    \centering
    \begin{subfigure}[b]{0.49\columnwidth}
        \centering
        \includegraphics[width=\linewidth]{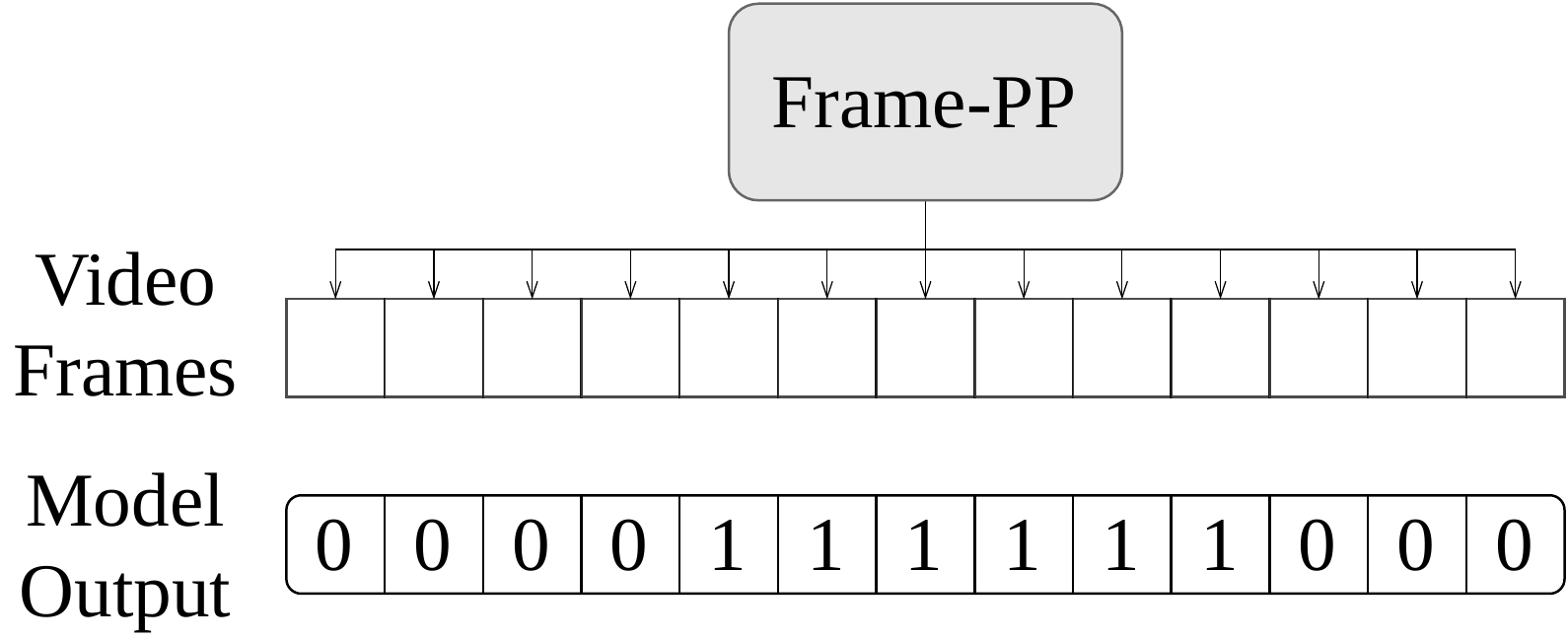} 
        \caption{
         \textbf{\framepp approach}}
        \label{fig:framepp}
    \end{subfigure}
    \hfill
    \begin{subfigure}[b]{0.49\columnwidth}
        \centering
        \includegraphics[width=\linewidth]{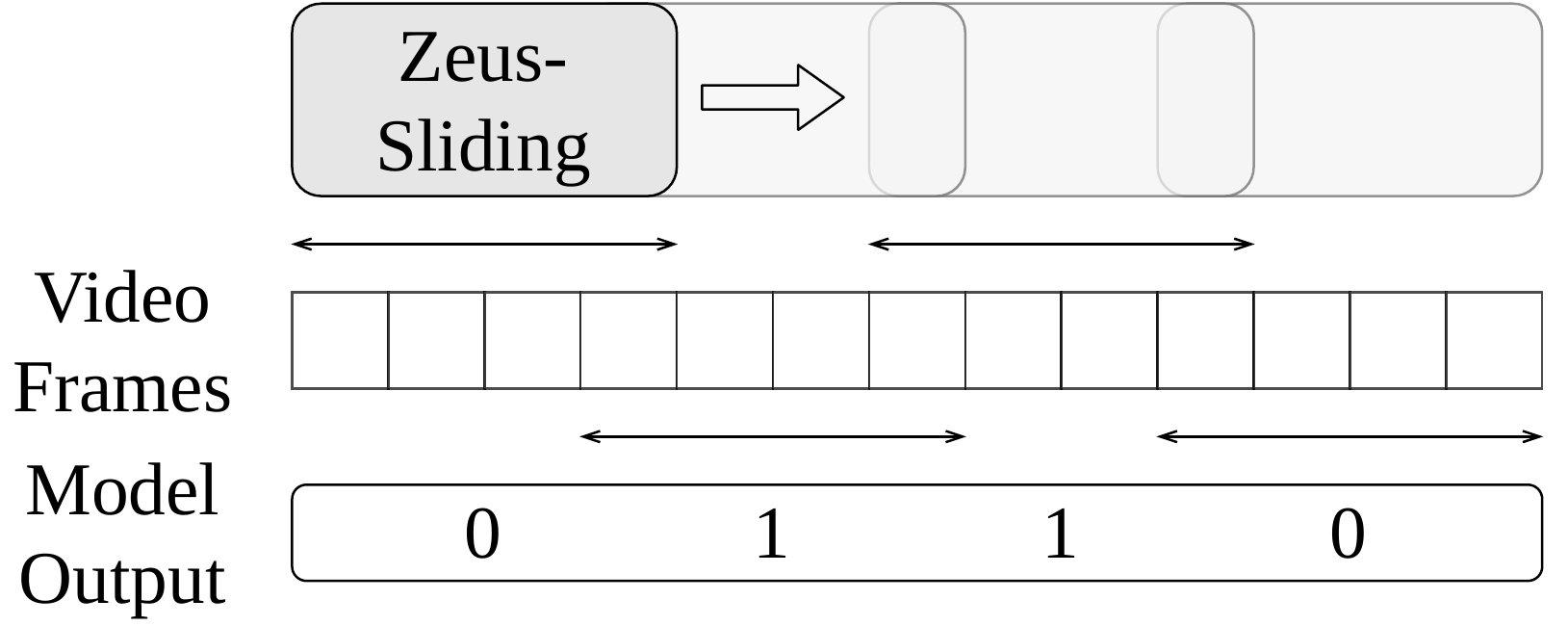} 
        \caption{
         \textbf{\windowpp approach}}
        \label{fig:sliding}
    \end{subfigure}
    \caption{ 
    	Frame- and window-based techniques for action localization in videos.
     }
\end{figure}

\PP{Frame-based filtering} 
Recent VDBMSs primarily focused on accelerating content-based
\textit{object detection} queries in
videos~\cite{kang2017noscope,kang13blazeit,lu2018accelerating}.
They rely on quickly filtering frames that are not likely to satisfy the
query's predicate using different \textit{proxy models}.
The VDBMS may use this frame-based filtering technique to process action queries
by filtering frames in the video that do not satisfy the action predicate.
We refer to this technique as frame-level probabilistic
predicate (\abbrev \framepp~\cite{lu2018accelerating}).
~\cref{fig:framepp} illustrates an example.

\PP{Window-based filtering} Recent solutions~\cite{simonyan2014two, tran2018closer} take a fixed-length segment in the video as input and return a binary label indicating the presence or
absence of an action.
The input segments contain three knobs:
(\resolution of each frame, \segmentlength -- number of frames in the segment, \samplingrate -- frequency at which each frame is sampled).
We refer to these as a \configuration tuple.
To process \textit{action queries}, a na\"ive solution that the VDBMS
could use is to apply the deep learning model
in a sliding window fashion on the video to detect actions and localize
action boundaries, as illustrated in~\cref{fig:sliding}.
The inputs to the model are video segments of a single \textit{fixed} configuration.
We refer to this baseline technique as \windowpp.
\rev{Another approach is to extend the frame-based filtering techniques in~\cite{kang2017noscope,kang13blazeit,lu2018accelerating} to segments
and build proxy models that can quickly filter irrelevant segments.
We refer to this baseline technique as \segmentpp.}

\PP{Challenges}
To efficiently process action queries, the VDBMS must handle three challenges:

\vspace{0.05in}\noindent\emph{\RomanNumeralCaps{1}. Task Complexity.}
Actions involve a complex interaction between objects within a video segment.
For example, the action \textit{Pole Vault} comprises of:
a person carrying a pole, running, and jumping over a hurdle using the pole.
Notice how the action comprises of multiple entities, and an interaction between these agents
in a specific manner.
The VDBMS has to detect all such actions and localize the start/end of the actions.
%
%
\rev{These queries must accurately capture the
temporal context and the scene complexity; this is too difficult for \framepp and \segmentpp
which apply lightweight frame- and segment- level filters.}
%

\vspace{0.05in}\noindent\emph{\RomanNumeralCaps{2}. Computational Complexity.}
\windowpp may localize complex actions using a deep neural network (\eg 
\rthreed~\cite{tran2018closer}).
But these networks are computationally expensive.
At a frame resolution of 720$\times$720, the \rthreed model runs at 2 frames per
second (fps) on a 16-core CPU, or at 13 fps on a server-grade GPU.
Further, both \framepp and \windowpp are sample inefficient - they require processing a large 
chunk of frames or (overlapping) video segments to achieve a desired accuracy.
%
%
%

\vspace{0.05in}\noindent\emph{\RomanNumeralCaps{3}. Accuracy Targets.}
While processing action queries, it is important to satisfy the
user-specified accuracy target, trading between accuracy and query performance.
\rev{\framepp and \segmentpp struggle to reach the target accuracy due to their inability
to capture the temporal context and scene complexity respectively.}
Consider the illustrative example in~\cref{fig:left-right-turn}.
None of the individual frames is sufficient \rev{for \framepp} to localize the action boundaries.
Meanwhile, each \windowpp model provides only a single
accuracy configuration throughout the video.


%




\PP{Our Approach}
We present \sys, a VDBMS designed for
efficiently processing action queries.
Unlike prior systems that rely on proxy models, \sys takes a novel approach based on
\textit{adaptive input segments}.
In particular, at each time step, it uses an Adaptive Proxy Feature Generator ({\tt APFG}),
a module that adaptively generates latent features of the video segment
at a small cost,
thus enabling (1) classification of the action class of the segment, and (2)
choosing the next input segment to process from a large space of possible
inputs.
We refer to these intermediate features as \feature{s}. 
%
%
\sys configures the input knobs when picking the next segment.
By doing so, \sys also quickly skims through the irrelevant
parts of the video. \looseness=-1
%

%
%
%
%
%
%
%

It is challenging to choose the optimal knob settings for deriving the next
segment at each time step.
Consider an agent that:
(1) uses slower knob settings if the current segment is an action, and 
(2) uses faster knob settings otherwise.
We refer to this baseline with hard-coded rules as \heuristic. 
With this technique, the VDBMS efficiently skims through non-action frames.
However, it does not have fine-grained control over the target accuracy, 
since the rules only indirectly affect accuracy.
Consider the case where the action spans the entirety of the video.
\heuristic would process the entire video with slower knobs, even when the 
accuracy target allows faster processing with a few mispredictions.
Further, \heuristic does not scale well to multiple knob settings.

We formulate choosing the optimal input segments at each time step as a
\textit{deep reinforcement learning} (\rl) problem~\cite{mnih2013playing}.
During query planning, \sys trains an \rl agent that learns to
pick the optimal knob settings at each time step using the proxy features.
To meet the accuracy target, the \opt incorporates novel
\textit{accuracy-aware aggregate rewards} into the training process.
%
%
%
The use of an \rl agent enables \sys to:
(1) maintain fine-grained control over accuracy, 
(2) achieve higher query throughput over the other baselines by
automatically using the optimal knob settings at each time step, and 
(3) scale well to multiple knobs.
We provide a qualitative comparison of the techniques for processing
action queries in ~\cref{tab:baselines-overview}. \looseness=-1


\begin{table}[t!]
    \begin{adjustbox}{width=\columnwidth,center}
    \begin{tabular}{cccccc}
    \toprule
                & \textbf{Sequence} & \textbf{Adaptive} & \textbf{Auto-Knob} &\textbf{ Accuracy}\\
                &  \textbf{Inputs} &   \textbf{Inputs}  & \textbf{ Selection} &  \textbf{Targets} \\ \midrule
    \framepp    & &    &    &  \\
    \rev{\segmentpp}  & \rev{\cmark} & & & \\
    \windowpp   &  \cmark &   & & \cmark\\
    \heuristic  &  \cmark & \cmark &   &  \\
    \sys        & \cmark & \cmark & \cmark & \cmark\\ \bottomrule
    \end{tabular}
    \end{adjustbox}
    \caption{
        \textbf{Techniques for processing action queries.}
        \sys 
        (1) operates on a sequence of frames at a time, 
        (2) adaptively selects input segments to boost performance, 
        (3) automatically chooses the knob settings, and 
        (4) has fine-grained control over query accuracy.
        }
    \label{tab:baselines-overview}
\end{table}

%
%

\PP{Contributions}
In summary, the key contributions are:
\squishitemize
\item \rev{We highlight the limitations of SOTA filtering techniques on action
queries(~\autoref{sec:introduction}).}
We present a novel approach for circumventing these limitations by  using proxy
features generated by an adaptive action recognition model
(~\autoref{sec:execution}).
\item We develop a reinforcement learning-based \opt that trains an \rl agent to
dynamically choose the \rev{optimal input configurations for the action
recognition model} (~\autoref{sec:planning}). 
\item We propose an accuracy-aware aggregate reward for training the
\rl agent and meeting the user-specified 
accuracy target(~\autoref{sec:aggregate-reward-allocation}).
\item We implement the \rl-based \opt and \proc in \sys.
We evaluate \sys on six action queries from three datasets and demonstrate that
it significantly improves over state-of-the-art in terms of query
processing, while consistently meeting the accuracy target
(~\autoref{sec:evaluation}).
\squishend


%% file: overview.tex
\section{Background}
\label{sec:action-localization}

%
%
%

%
Action queries focus on
(1) action classification~\cite{karpathy2014large, carreira2017quo,
tran2018closer} which answers  \textit{what} actions are present in a long,
untrimmed video, and
(2) temporal boundary localization~\cite{shou2016temporal, chao2018rethinking} for \textit{where} these actions are present in
the video.
We refer to the combined task of detecting, and localizing actions in videos as action localization (AL).

AL focuses on retrieving events in a video that happen over a span of time, and
that often involve interaction between multiple objects.
%
%
For example, \rev{a} traffic analyst may be interested in examining video segments
that contain pedestrians crossing the road from left to right (\crossright) from a
corpus of videos obtained from multiple cameras.
This query requires detection of:
(1) pedestrians in the frame,
(2) \textit{walking} action of the pedestrian, and
(3) trajectory of the person (left to right).
The VDBMS must also accurately detect the start and
end of this action.
%




\begin{figure}
    \centering
    \begin{subfigure}[b]{0.44\columnwidth}
        \centering
        \includegraphics[width=\linewidth]{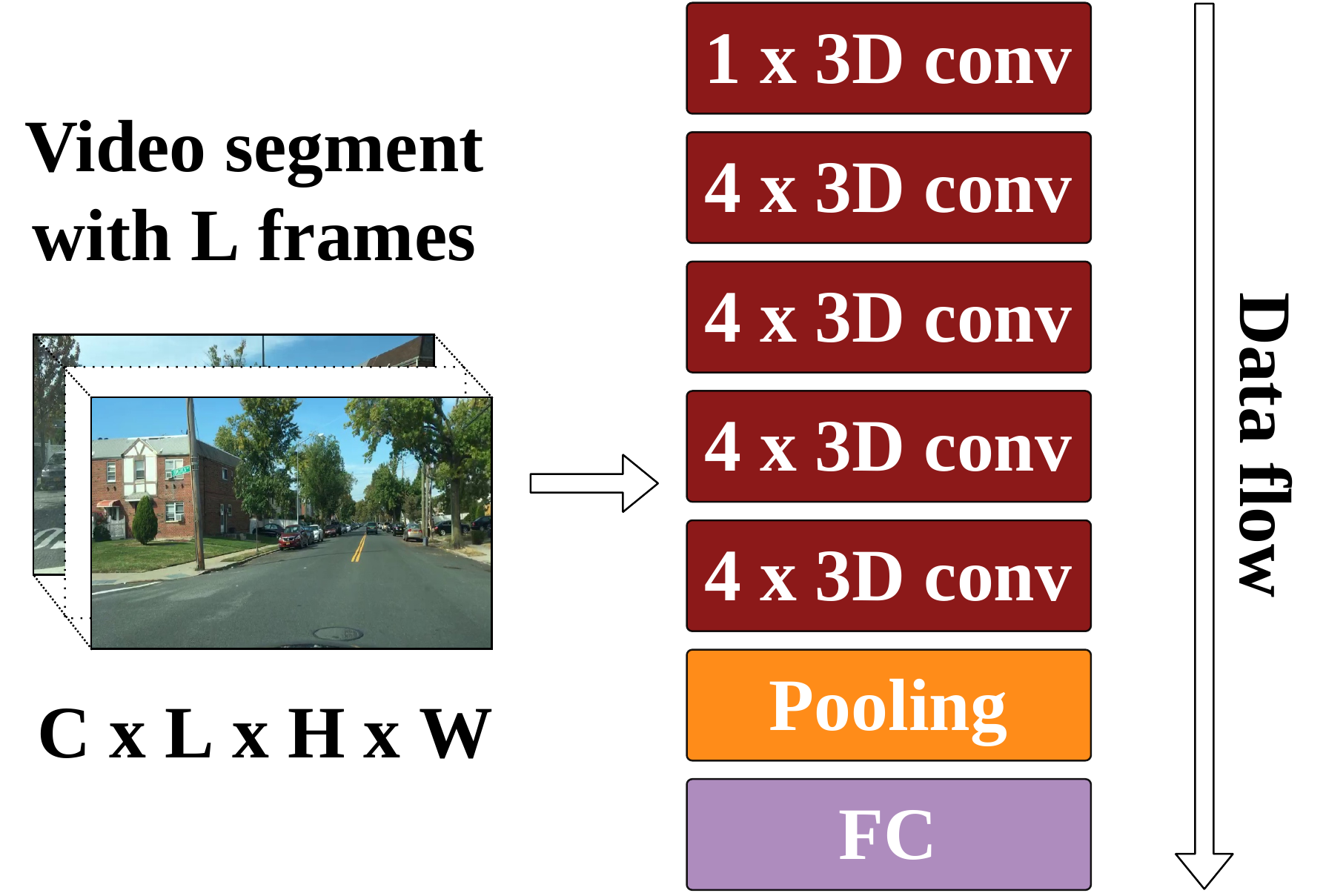}
        \caption{
         \textbf{3D convolutions in \rthreed}}
        \label{fig:r3d-conv}
    \end{subfigure}
    \hfill
    \begin{subfigure}[b]{0.54\columnwidth}
        \centering
        \renewcommand{\arraystretch}{1.2}
        \resizebox{\textwidth}{!}{%
        \begin{tabular}{cc}
            \hline
            Kernel             & Output size              \\ \hline
            \{3$\times$7$\times$7, 64\} $\times$ 1  & 64 $\times$ L $\times$ H/2 $\times$ W/2       \\ \hline
            \{3$\times$3$\times$3, 64\} $\times$ 4  & 64 $\times$ L$ \times$ H/2$ \times$ W/2      \\ \hline
            \{3$\times$3$\times$3, 128\} $\times$ 4 & 128 $\times$ L/2 $\times$ H/4 $\times$ W/4   \\ \hline
            \{3$\times$3$\times$3, 256\} $\times$ 4 & 256 $\times$ L/4 $\times$ H/8 $\times$ W/8   \\ \hline
            \{3$\times$3$\times$3, 512\} $\times$ 4 & 512 $\times$ L/8 $\times$ H/16 $\times$ W/16 \\ \hline
            1$\times$1$\times$1 & 512 $\times$ 1 \\ \hline
            512 $\times$ $n\_classes$ & $n\_classes$ $\times$1 \\ \hline
        \end{tabular}
        }
        \caption{
         \textbf{Data flow in \rthreed}}
        \label{fig:r3d-data-flow}
    \end{subfigure}
       \caption{
            The \rthreed network consists of 17 3D convolutional layers followed by adaptive average pooling
            and fully-connected layers.
            The input to the network is a video segment of length L frames.
            In successive 3D convolutional operations, the input is reduced to eventually get a 512
            dimensional feature vector.
        }
       \label{fig:r3d}
\end{figure}

\PP{Neural Networks for AL}
Researchers have proposed deep neural networks (DNNs) for AL.
An early effort focused on using 2D convolutional neural networks (CNNs) to
classify actions~\cite{karpathy2014large}.
2D-CNNs are typically used for image classification and object
detection and do not effectively capture information along the
temporal domain (\ie across frames in a segment).
%
%
For the same reason, the frame-level filtering in recently proposed
VDBMSs~\cite{kang2017noscope,lu2018accelerating,kang13blazeit} is
suboptimal for AL.
%

%
%
%
%
Frame-level filtering is also unable to accurately detect the action boundaries.
This is because frames before, during, and after the scene of the action can be visually indistinguishable.
In our evaluation section, we show that \framepp operates at prohibitively low
accuracies for action queries (\autoref{sec:eval-endtoend}).
%
%
%
%

\PP{3D-CNNs for AL}
3D-CNNs (\eg \rthreed~\cite{tran2018closer}) apply the convolution operation along both spatial and temporal
dimensions (~\cref{fig:r3d-conv}).
This allows them to aggregate information along the temporal domain.
More formally, consider an input video segment of size
$C \times L \times H \times W$, where
$C$ is the number of channels,
$L$ is the number of frames in the segment, and
$H$ and $W$ are the height and width of each frame.
%
For each channel in $C$, the $ L \times H \times W $ cuboid is convolved with a
fixed size kernel to generate a 3D output.
Outputs for each channel are pooled to produce a single 3D output in $ L' \times H' \times W' $ that corresponds to one kernel.
With $K$ kernels, we obtain a $ K \times L' \times H' \times W' $ matrix 
as the output of each layer.
The overall network consists of a series of such spatio-temporal 3D
convolutional blocks.
%

\begin{figure}[t!]
    \includegraphics[width=0.95\linewidth]{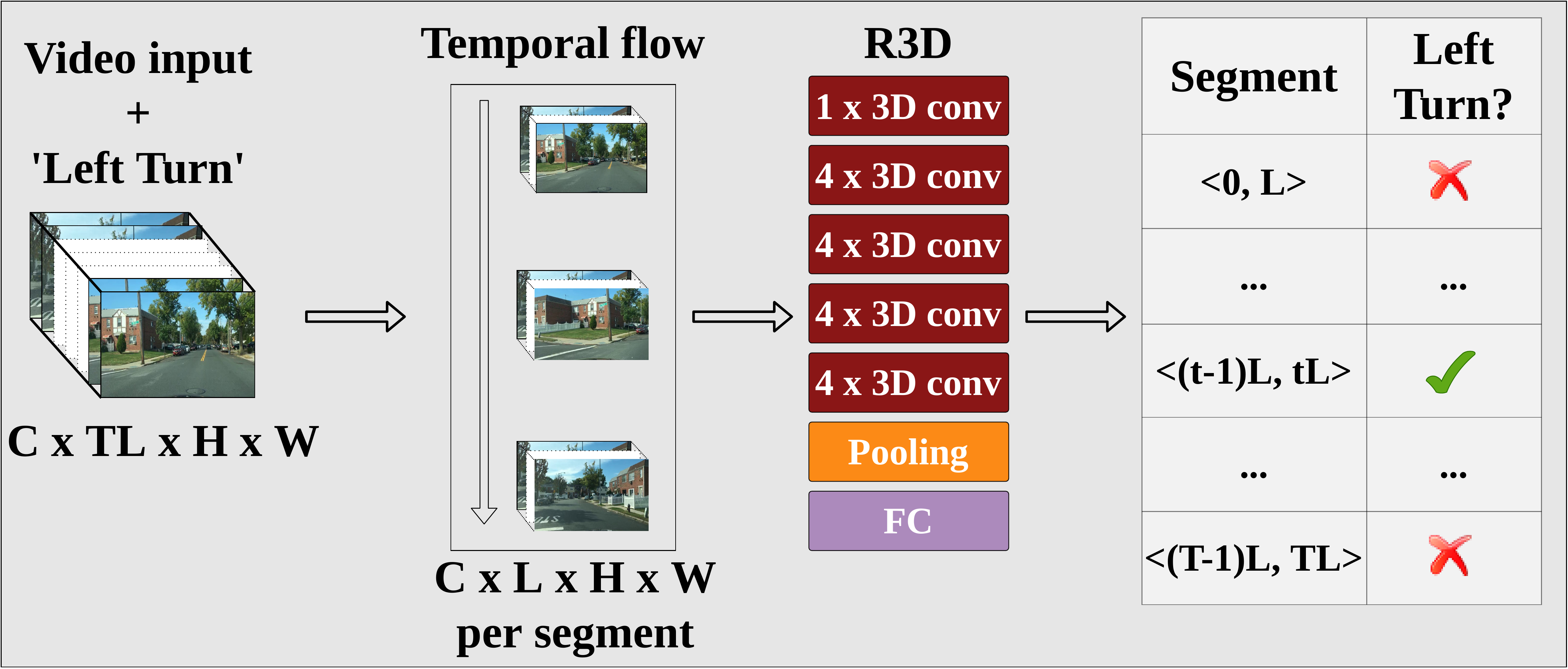}
    \caption{ 
    In \windowpp, the input video of T $\times$ L frames is divided into T segments
    of L frames each.
    Each segment is then processed by the \rthreed network to generate
    binary class labels to determine the presence/absence of action.
    }%
    \label{fig:baseline}
\end{figure}

%

\PP{\windowpp}
As demonstrated in~\cref{fig:baseline}, the \windowpp technique (recall~\autoref{sec:introduction}) uses the \rthreed network
in a sliding window fashion over the input video.
For each segment of size $L$,  all frames are stacked together to form a 4D
input segment and are passed to the \rthreed network, which generates class labels for the current window;   the VDBMS slides
the window forward by a specified number of interval frames.
In this manner, the network classifies and localizes the actions.

%
%
However, \windowpp is computationally expensive.
The \rthreed model has significantly more parameters -- 33.4 million which is 3$\times$
higher than the corresponding 2D resnet-18 model.
On an NVIDIA GeForce RTX 2080 Ti GPU, at the resolution of 480x480, the \rthreed
network processes 27 frames per second (fps), 
while its 2D counterpart processes frames at 156 fps. 
This limits current VDBMSs to analyse video data at scale.

%
%
%
%
%


\subsection{Problem Formulation}
\label{sec:overview-workflow}
Given an input video, the objective of our system, \sys, is to efficiently
process the video and find segments that contain an action.
%
%
\vspace{0.05in}Each video may contain different types of actions.
Given a \textit{label function} $\mathbb{L}(n)$ that provides the \textit{oracle} action
label at frame $n$, the binary label function for an action class X is defined as:
\begin{equation}
f_X(n) = \begin{cases}
             1  & \text{if }\, \rev{\mathbb{L}(n) = X} \\
             0  & \text{if }\, \rev{\mathbb{L}(n) \neq X}
       \end{cases} \quad
       \label{eqn:label-function}
\end{equation}
\sys measures the accuracy of the query with respect to the oracle label
function \rev{$\mathbb{L}(n)$.
A binary ground truth label for a segment is generated using intersection-over-union (IoU) over the frame-level ground truth labels.
A given segment of length K frames is labeled as a true positive if IoU>0.5 over labels $\mathbb{L}(n)$ to $\mathbb{L}(n+K)$.
\sys compares the output of the classifier and the
segment index against this segment label to compute the query accuracy.
}
\sys seeks to efficiently process this AL query while meeting the target
accuracy, \ie the accuracy target specified by the user with the query.

%


%% file: execution.tex
\begin{figure}[t]
    \centering
    \includegraphics[width=0.95\linewidth]{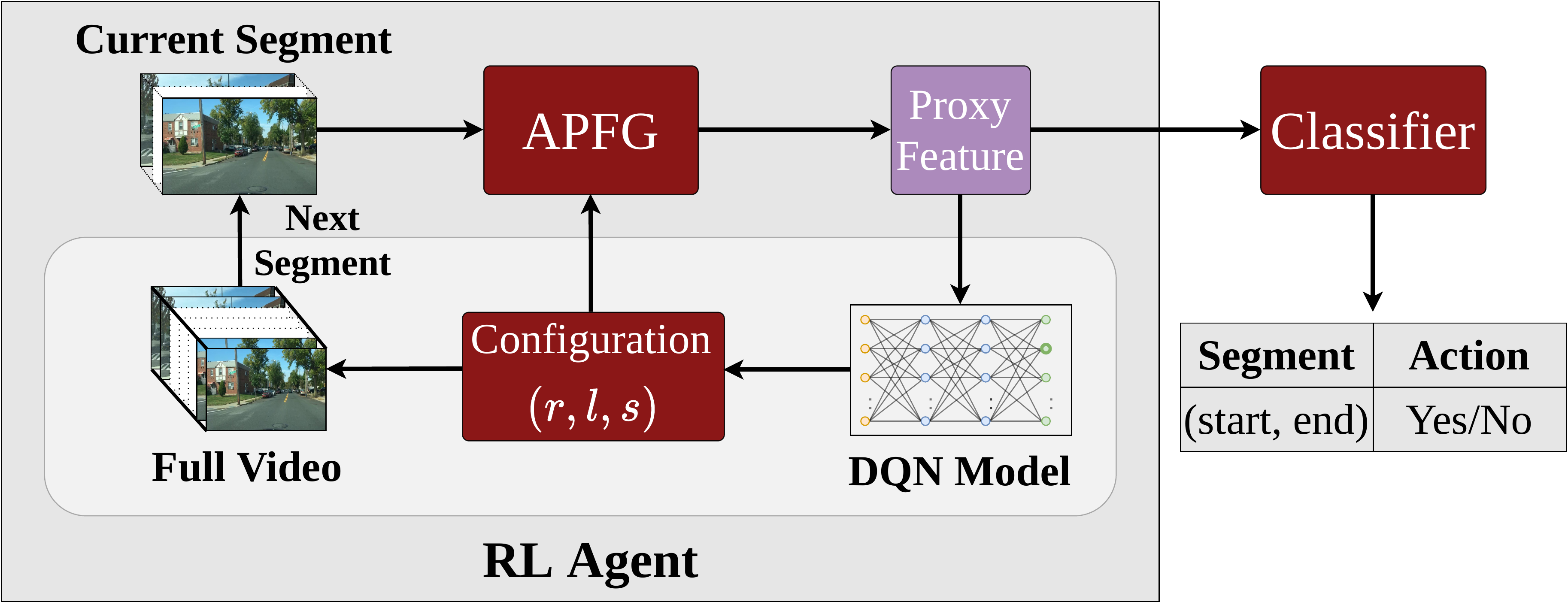}
    \caption{
     \textbf{Architecture of \proc in \sys --}
     The \proc uses a deep \rl agent to process action queries.
     It takes as input the \feature generated by the \apfg, to
     output a \configuration, which is used to construct the next input segment.
     The \feature is also processed by a classifier to predict the
     presence of action in the current segment.    
     }
    \label{fig:system}
\end{figure}

\begin{figure*}[t!]
    \tiny
    \centering
    \resizebox{\textwidth}{!}{%
    \begin{tabular}{cc @{\hspace{0.2\tabcolsep}} c @{\hspace{0.2\tabcolsep}}c @{\hspace{0.2\tabcolsep}} c @{\hspace{0.2\tabcolsep}} c}
    &
    \includegraphics[width=0.1\linewidth]{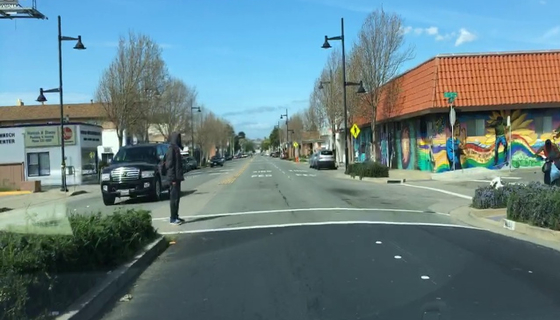} &
    \includegraphics[width=0.1\linewidth]{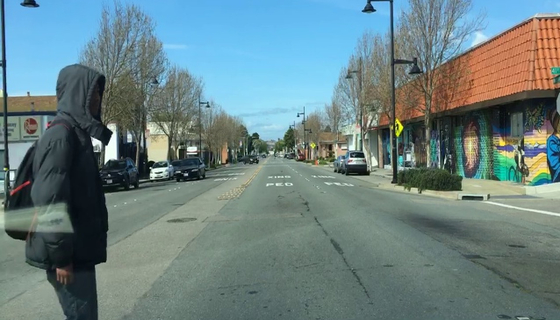} &
    \includegraphics[width=0.1\linewidth]{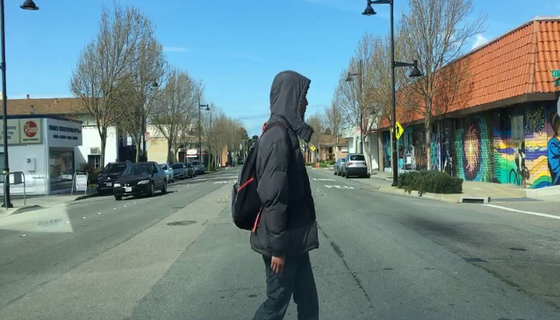} &
    \includegraphics[width=0.1\linewidth]{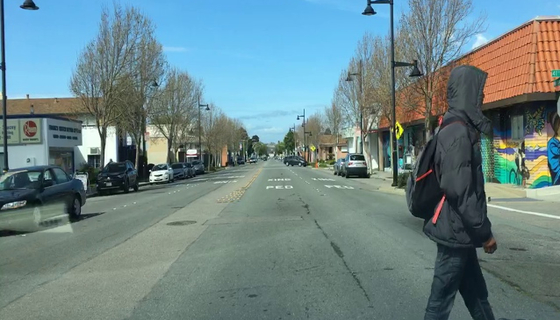} &
    \includegraphics[width=0.1\linewidth]{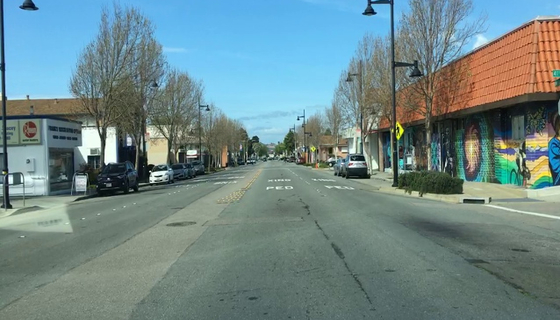}\\ \hline
    
    \textbf{Time Step} & t=1 & t=2 & t=3 & t=4 & t=5 \\ \hline
    Frame Number & f=1 & f=65 & f=97 & f=101 & f=113 \\ 
    \configuration & (150, 8, 8) & (200, 6, 4) & (300, 4, 1) & (250, 6, 2) & (150, 8, 8) \\ 
    Segment & (1, 64) & (65, 96) & (97, 100) & (101, 112) & (113, 176)\\ \hline
    \textbf{\apfg Prediction} & \texttt{NO-ACTION} & \texttt{ACTION} &
    \texttt{ACTION} & \texttt{ACTION} & \texttt{NO-ACTION}\\ \hline
    \end{tabular}
    }
    \caption{
        \textbf{Illustration of \sys on \crossright action --}
        At each time step, \sys picks from a set of configurations to generate
        the next input to the \apfg.
        Each \configuration is a 3-tuple (\resolution, \segmentlength, \samplingrate).
        %
    }
    \label{fig:example}
\end{figure*}

\section{Executing Action Queries}
\label{sec:execution}

%
%
%

\PP{Overview}
We propose a system, \sys, to efficiently answer AL queries. 
\cref{fig:system} demonstrates the architecture of the \proc in \sys with the following components:

\squishitemize
\item Segment: Each video segment is
characterized by <{\tt start}, {\tt end}> which represent the start and end
frames of the action.
{\tt start}, {\tt end} $\in[1, N]$, where $N$ is the total number of frames. 
\item \feature: A fixed-size vector of floats that succinctly represents an
input video segment.
\item \apfg: Denotes an Adaptive Proxy Feature Generator.
It is a collection of action recognition models that \emph{adaptively} generate \feature{s} based
on the configuration (\eg a collection of \rthreed models that operate on input
segments of varying segment lengths and resolutions).
\item Classifier: A model that emits the action label.
\item \configuration $(r,l,s)$ - Concrete setting of three knobs -
\resolution, \segmentlength, and \samplingrate (\autoref{sec:introduction}).
\squishend 

%
%
\vspace{0.05in}
At a given time step t, \sys takes a video segment of length $l_t$,
resolution $r_t \rightarrow (h_t, w_t)$, sampled once every $s_t$ frames, to generate
a 4D vector of size $3 \times l_t \times h_t \times w_t$ as the input.
%
The \apfg module processes this input and generates a proxy feature vector $\hat{Z}_{t}$,
which is then passed to a classification network that predicts the presence or absence
of an action in the current segment.
%
%
The \rl agent also uses $\hat{Z}_{t}$ to generate the next \configuration
$(r_{t+1}, l_{t+1}, s_{t+1})$, using which the \apfg constructs the next segment of size
$ 3 \times l_{t+1} \times h_{t+1} \times w_{t+1} $ sampled at a frequency
of $s_{t+1}$. 
%
%
%

\PP{\apfg}
Unlike prior VDBMSs, \sys relies on \feature{s} as opposed to a proxy model.
In particular, it uses the \apfg to generate the \feature{s} for the segments
in the video.
The \apfg is an ensemble of action recognition models (\eg \rthreed) that can
adaptively process input segments of varying resolutions and segment lengths
to generate the \feature{s}.
\rev{More specifically, the ensemble contains models trained for different
resolutions and segment lengths.
Typically, models trained with one configuration have a lower accuracy when
tested on other configurations.
So, the ensemble design provides a better accuracy and flexibility compared
to a single \rthreed model, that only supports segments of fixed resolution and length.}
Moreover, the \apfg can leverage a wide range other of action recognition models.
So, the \rthreed model can be easily replaced with better
(faster, more accurate) models if necessary.
\rev{Since training models for each configuration is expensive,
we present an optimization to lower this cost in~\autoref{sec:implementation}.} 

\sys processes the \feature{s} from \apfg using a classification network to determine
the presence or absence of actions.
This technique allows \sys to infer complex relationships present in the scene.
Besides classifying the given scene, \sys uses the \feature to also
construct the next input to the \apfg.
In particular, it trains a deep \rl agent to tune the configuration with 3 knobs.
The agent learns to use segments of low-resolution, coarse-grained sampling (fast
\configuration) for portions of the video that are unlikely to contain an action.
Otherwise, it uses high-resolution, fine-grained sampling (slow \configuration).
%

\PP{Illustrative example}
Consider the example shown in \cref{fig:example}.
\sys localizes a \crossright action (\ie a pedestrian crossing the street
from left to right).
%

At time step $t=1$, \sys starts at frame number 1.
Assume \sys has access to a \feature $\hat{Z}_{0}$ from the previous time
step.
Since there is no \crossright in this segment,
\sys uses a low resolution $150\times150$, high segment length $(8)$ and high
sampling rate $(8)$ to construct the input to the \apfg.
Henceforth, the knob settings will be represented as a \configuration - $(150, 8, 8)$ in this case.
\rev{We use the most accurate (and computationally expensive) configuration as
this first configuration at time step 1.
We only use square-shaped frames with equal height and width.}
The \apfg processes the segment $(1, 64)$ sampled once every 8 frames to generate a \feature
$\hat{Z}_{1}$ and a prediction \texttt{NO-ACTION}.
Thus, \sys processes $8 \times 8$ frames in this time step at a low resolution of $150$,
jumping to the frame $65$.
%
%
At $t=2$, \sys processes the \feature $\hat{Z}_{1}$ from $t=1$.
Since $\hat{Z}_{1}$ aggregates information for frames 1 to 64,
the agent learns the possibility of \crossright at the end of the segment (f = 64).
So, \sys uses a slower \configuration of $(250, 6, 4)$ to generate $\hat{Z}_{2}$ and
the prediction \texttt{ACTION}.
Notice that the \configuration change from $t=1$ to $t=2$ is \textit{gradual},
since the possibility of \crossright has not been fully established in $t=1$.
At $t=3$, \sys processes the \feature $\hat{Z}_{2}$, 
which was generated from a higher resolution finely-sampled input.
So, \sys recognizes the presence of an action and uses a slow \configuration
of $(300, 4, 1)$ to generate $\hat{Z}_{3}$ and the prediction \texttt{ACTION}.
%
At $t=4$, \sys processes $\hat{Z}_{3}$ and generates the \configuration $(250, 6, 2)$.
It slightly increases the \configuration speed since it recognizes that
\crossright may end soon at f=100.
However, the prediction for this segment is still \texttt{ACTION}.
Finally, at $t=5$, \sys detects the end of \crossright and reverts
to the fastest \configuration of $(150, 8, 8)$.

Across five time steps, \sys performs 5 invocations of the \apfg, processes $176$
frames, with 160 of those frames processed at less than half of the highest resolution, 
while still accurately localizing the action boundaries from $(65, 112)$.
In this way, \sys efficiently and accurately localizes and classifies the
target action.

%% file: planning.tex
\section{Planning Action Queries}
\label{sec:planning}

We present the internals of the \opt in this section.
We begin with a discussion on how we formulate the RL problem
in~\autoref{sec:rl-formulation}. 
We then describe how the \opt picks the configurations that can be used for
processing a given query in~\autoref{sec:cost-estimation}.
In~\autoref{sec:dqn}, we present an overview of the RL algorithm used in
\sys.
We then describe the local reward function in~\autoref{sec:local-reward}.
Since this function does not always meet the accuracy constraint, we discuss a
reward aggregation strategy in~\autoref{sec:aggregate-reward-allocation}.
Lastly, in~\autoref{sec:accuracy-based-reward}, we outline how the \opt trains
an accuracy-aware \rl agent.
 
\subsection{RL Formulation}
\label{sec:rl-formulation} 
The goal of the agent is to choose configurations at each time step such that
the overall query processing time is minimized while the target accuracy is
satisfied (as discussed in~\autoref{sec:execution}).

\PP{Heuristics vs \rl}
In theory, \sys could leverage an agent that uses hard-coded rules
to determine the parameters for constructing the segment.
For example, consider the rule: if the video contains an \texttt{ACTION}
at the current time step, then the agent should use a lower sampling rate.
However, such a hard-coded low sampling rate is sub-optimal since the agent
does not have fine-grained control over the query accuracy.
\rev{In \cref{fig:example}, \heuristic would use the slow \configuration $(300, 4,
1)$ at time step $t=3$ and get the \texttt{ACTION} signal from the \apfg.
So, \heuristic again uses the slow configuration, and only processes 4 frames in
time step $t=4$ as opposed to \sys that processes 12 frames in time step $4$ (as shown in~\cref{fig:example}).
\heuristic keeps using the slow configuration for time steps $5$, $6$, and $7$,
since the prediction at each of these time-steps is \texttt{ACTION}.
Effectively, \heuristic reaches frame 112 only at time step $7$, while \sys does so by
time step $4$.}
This would be sub-optimal since the agent would process 16 frames in the
sequence (97, 112) at a significantly lower throughput, while
potentially overshooting the accuracy target.
So, \heuristic cannot trade-off the accuracy for throughput.
Furthermore, the number of such rules will increase with the number of
available configurations and may interact in unexpected
ways~\cite{10.1145/66926.66962}.

To circumvent these problems, \sys uses an \rl-based agent that adaptively
changes the segment's parameters.
For a given accuracy target, \sys trains the \rl agent to trade-off the
excess accuracy for throughput improvement.
We compare the performance of \sys against \heuristic in \autoref{sec:eval-endtoend}.

\PP{Markov Decision Process}
Most RL problems are modelled as Markov Decision Processes
(MDP)~\cite{howard1960dynamic, shoham2003multi}.
An RL problem formulated as MDP is characterized by an \textit{agent}
that takes \textit{actions} in an \textit{environment}.
Each action changes the \textit{state} of the agent in the environment,
and the agent receives a \textit{reward} for its actions.
We formulate our problem as an MDP with the following components:

\squishitemize
\item \textbf{Environment:} The set of training videos V \{$v_1, v_2, \cdots, v_N$\}.
\item \textbf{State:} 4D tensor for the segment $(s, e)$.
\item \textbf{Reward:} Reward function for training the agent (\autoref{sec:local-reward})
that maps each step taken by the agent to a scalar reward value.
\item \textbf{Configuration:} \configuration chosen by the agent at each time step from
the set of available configurations $\{c_1, c_2, \cdots, c_n\}$.
\squishend

\begin{table}[t]
    \centering
    \begin{adjustbox}{width=\columnwidth,center}
    \begin{tabular}{ccc|cc}
    \toprule
    \textbf{Resolution} & \textbf{Segment Length} & \textbf{Sampling Rate} &
    \textbf{Throughput (fps)} & \textbf{F1-score} \\\midrule
    150 & 4 & 8 & 1282  & 0.57 \\
    200 & 4 & 4 & 553   & 0.82  \\
    250 & 6 & 2 & 285   & 0.86  \\
    300 & 6 & 1 & 115   & 0.91  \\ \bottomrule
    \end{tabular}
    \end{adjustbox}
    \caption{
        \textbf{Illustrative list of configurations used by \sys --}
        Each configuration has associated throughput and accuracy metrics.
        %
        %
    }
    \label{tab:example-configs}
\end{table}

\subsection{Configuration Planning}
\label{sec:cost-estimation}
Given an input query with a target accuracy, the \opt seeks to maximize
throughput while satisfying the accuracy constraint.
To achieve this goal, it first collects the appropriate settings for all of the
knobs for the given query (\ie a \configuration).
Each configuration has two associated \textit{cost metrics}:
(1) throughput (fps), and (2) accuracy.
For example, using coarse-grained sampling leads to faster (but less accurate)
query processing.

\PP{Pre-Processing}
In a one-time pre-processing step  during the query planning phase,
\sys computes the cost metrics associated with each configuration on a held-out
validation dataset.
It uses \windowpp (\autoref{fig:baseline}) to compute these metrics.
\cref{tab:example-configs} lists a few configurations for the \crossright query
after this step.
Notice that the throughput and the accuracy of the configurations is inversely
proportional.
This pattern allows the VDBMS to trade-off throughput and accuracy by choosing
faster configurations if the accuracy is greater than the
accuracy target.
%

%
%



\subsection{Deep Q-Learning}
\label{sec:dqn}

\begin{algorithm}[t]
\SetArgSty{textnormal}
\footnotesize
\setstretch{1.2}
  \SetKwInOut{Input}{Input}
  \SetKwInOut{Output}{Output}
  \SetKwInOut{Initialize}{Initialize}
  \Input{Query Q, Number of training episodes = T\\
        Set of N training videos V = \{$v_1, v_2, \cdots, v_N$\}\\}
  \Output{Trained DQN-Network $\phi$}
  \Initialize{
        Ground\_Truths(GT); APFG(U)\; \\ \label{alg:dqn:start-init-algo}
        Replay\_Buffer(B); DQN\_Network($\phi$)\; \\
         \label{alg:dqn:end-init-algo}
  }
  %
        \For{i $\gets 1$ to T}{
            \tcp{Initialize episode}
            V\textsubscript{i}, GT\textsubscript{i} $\gets$ Random\_Order(V, GT)\; \label{alg:dqn:start-init-eps}
            \texttt{segment\textsubscript{curr}} $\gets$ Init\_Segment(V\textsubscript{i})\;
            \texttt{state\textsubscript{curr}} $\gets$ U(\texttt{segment\textsubscript{curr}})\; \label{alg:dqn:end-init-eps}
            \While{\text{$idx$(\texttt{segment\textsubscript{curr}}) < size(V)}}{
                \tcp{Generate experiences}
                \texttt{config\textsubscript{curr}} $\gets$ argmax($\phi$(\texttt{segment\textsubscript{curr}}))\; \label{alg:dqn:start-gen-exp}
                \texttt{reward},  \texttt{segment\textsubscript{next}} $\gets$ TraverseVideo(V\textsubscript{i}, GT\textsubscript{i}, \texttt{config\textsubscript{curr}})\; \label{alg:dqn:gen-exp-reward}
                \texttt{state\textsubscript{next}} $\gets$ U(\texttt{segment\textsubscript{next}}, \texttt{config\textsubscript{curr}})\; \label{alg:dqn:gen-exp-udf}
                \tcp{push experience to replay buffer}
                B.push(\texttt{state\textsubscript{curr}}, \texttt{config\textsubscript{curr}}, \texttt{reward}, \texttt{state\textsubscript{next}})\; \label{alg:dqn:replay-push}
                \tcp{Update DQN model}
                \uIf{$update_{\phi}$}{
                    \texttt{minibatch} $\gets$ B.sample(\texttt{batch_size})\; \label{alg:dqn:start-update}
                    \texttt{q_values}, \texttt{targets} $\gets$ $\phi$(\texttt{minibatch})\; 
                    \texttt{loss} $\gets$ HuberLoss(\texttt{q_values}, \texttt{targets})\;
                    update\_weights($\phi$, \texttt{loss})\; \label{alg:dqn:end-update}
                }
                \texttt{state\textsubscript{curr}} $\gets$ \texttt{state\textsubscript{next}}\;
                \texttt{segment\textsubscript{curr}} $\gets$ \texttt{segment\textsubscript{next}}\;
            }
        }
  %
  \caption{Algorithm for training the \rl agent} 
  \label{alg:dqn}
\end{algorithm}

We now provide an overview of the RL algorithm used in \sys.
Deep Q-learning (DQN) is a variant of the Q-learning algorithm for
\rl~\cite{mnih2013playing}.
The original Q-learning algorithm takes numerical states as inputs and uses a
memoization table (called the Q-table) to learn the mapping of a given state
to a particular configuration.
In DQN, this mapping is learnt by a DNN that approximates the Q-table.
DQN works well when the state space of the inputs and/or the configuration space is large.
Since visual data is high dimensional, DQN is often used to train \rl agents that
operate on images or videos (\eg Atari games).

\PP{Training Process}
\cref{alg:dqn} outlines the DQN technique used by \sys to train an \rl agent for
action queries.
The training algorithm assumes the presence of ground truth labels,
and trained \apfg.
The DQN algorithm has two main components:
(1) the agent collects experiences by traversing through the video
(\cref{alg:dqn:start-gen-exp} to \cref{alg:dqn:replay-push}), and
(2) it updates the model parameters (model's mapping of the video and
configurations) using the collected experiences (\cref{alg:dqn:start-update} to
\cref{alg:dqn:end-update}). 
To generate a new experience, the agent applies the configuration \texttt{config\textsubscript{curr}}
output by the network (\cref{alg:dqn:start-gen-exp}) to transition from the current state
\texttt{state\textsubscript{curr}} to the next state \texttt{state\textsubscript{next}} in the video
(\cref{alg:dqn:gen-exp-reward} to \cref{alg:dqn:gen-exp-udf}).
The agent receives a \texttt{reward} for this transition (\cref{alg:dqn:gen-exp-reward}) (\autoref{sec:local-reward}).
This 4-tuple of (\texttt{state\textsubscript{curr}}, \texttt{config\textsubscript{curr}}, \texttt{reward}, \texttt{state\textsubscript{next}}) is called an
\textit{experience tuple}.
The agent continuously collects new experience tuples by traversing through the video, and
updates the model parameters periodically to account for the newly observed experience tuples.

One way to achieve this behavior is to always use the most recently generated experience tuples
to update the Q-network.
However, such tuples are from nearby segments in the video and
hence highly correlated.
DNN training process is known to be ineffective when learning with
\textit{correlated tuples}.
To overcome this problem, DQN uses an \textit{experience replay buffer}.
This is a cyclic memory buffer that stores the experience tuples from the last
\texttt{K} transitions (a hyperparameter).
While generating experiences, the agent pushes the new experience tuples
to the replay buffer (\cref{alg:dqn:replay-push}).
During the update step of DQN, \sys samples a mini-batch of experiences from the
replay buffer (\cref{alg:dqn:start-update}) and updates the model parameters (\cref{alg:dqn:end-update}).
This technique improves the model's \textit{sample efficiency} by reducing the
correlation between samples during a single update step. \looseness=-1

\PP{Convergence Speed}
DQN is notoriously slow to converge on large state spaces~\cite{mnih2013playing}.
It requires a large number of experience tuples to approximate the Q-value
function.
This problem is more significant for video segment inputs since 
experience generation is slower with 4D tensors.
Additionally, the replay buffer size is prohibitively large when the states are
raw 4D tensors.
Thus, representing states using raw 4D tensors is impractical even on
server-grade GPUs.
To circumvent this problem, \sys first generates the video segment \texttt{segment\textsubscript{next}}
during the transition (\cref{alg:dqn:gen-exp-reward}).
The \apfg then processes \texttt{segment\textsubscript{next}} to generate the
proxy features that serve as the next state \texttt{state\textsubscript{next}}
of the agent (\cref{alg:dqn:gen-exp-udf}).  
The \apfg is trained independently of the \rl agent
and its weights are \textit{frozen} during the RL training process.
This \feature is then used as the state for the \rl agent.
%

\subsection{Local Reward Allocation}
\label{sec:local-reward}

The agent's goal is to select a \configuration at each time step that
minimizes query processing time while satisfying the accuracy requirement.
%
%
To minimize processing time, the agent should skip more frames and pick
lower resolutions.
To meet the accuracy requirement, the agent should use high resolution and
fine-grained sampling in the vicinity of action segments within the video.
%


\begin{figure}
    \centering
    \begin{subfigure}[b]{0.32\columnwidth}
        \centering
        \includegraphics[width=\linewidth]{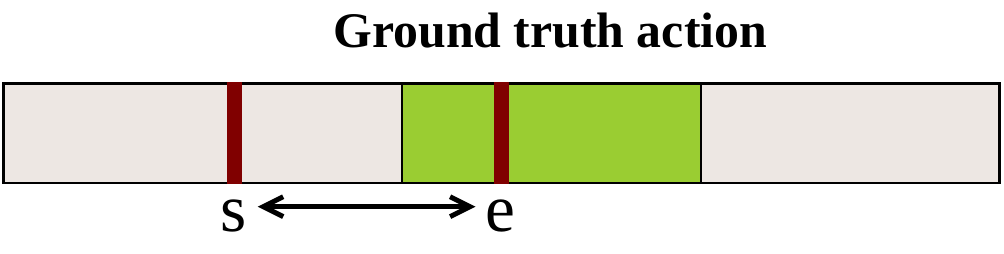}
        \caption{
         \textbf{Penalty}}
        \label{fig:penalty}
    \end{subfigure}
    \hfill
    \begin{subfigure}[b]{0.32\columnwidth}
        \centering
        \includegraphics[width=\linewidth]{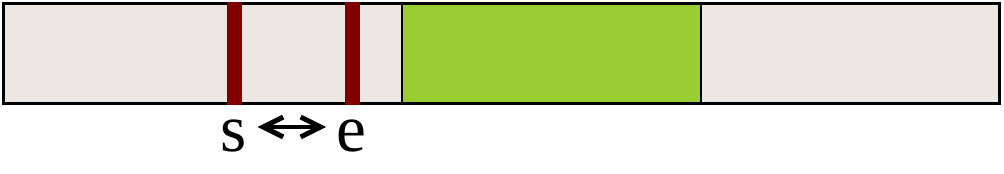}
        \caption{
         \textbf{Low reward}}
        \label{fig:lowreward}
    \end{subfigure}
    \hfill
    \begin{subfigure}[b]{0.32\columnwidth}
        \centering
        \includegraphics[width=\linewidth]{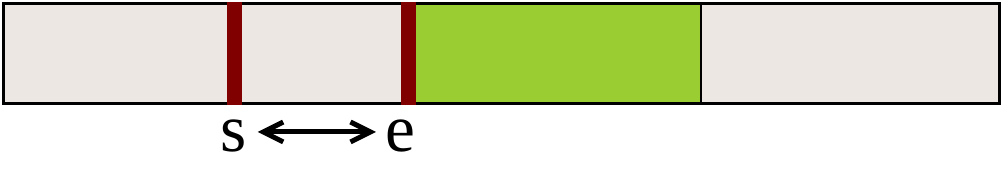}
        \caption{
         \textbf{High reward}}
        \label{fig:highreward}
    \end{subfigure}
       \caption{
            \textbf{Reward Scenarios --}
            The reward function (\autoref{sec:local-reward}) assigns reward to the
            agent differently for each of the scenarios shown above.
            %
        }
       \label{fig:rewardassignment}
\end{figure}

\PP{Local Reward Function}
To achieve these goals, one way to model the reward function (\cref{alg:dqn:gen-exp-reward}
in \cref{alg:dqn}) for the agent would be:
(1) reward decisions that increase the processing speed, and
(2) penalize decisions that lower accuracy.
At a given time step $t$, let us assume that the agent outputs the
\configuration - $(r, l, s)$.
With this configuration, the segment window that is processed in this time step
is $(s, e) \rightarrow (f_{curr}, f_{curr} + s*l)$, where $f_{curr}$ is the current
location of the agent.
Lets assume that each configuration $c$ has an associated scalar value $\alpha_{c}$
that represents the \textit{fastness} of a configuration (\ie throughput in~\cref{tab:example-configs}).
The faster the configuration, the greater the value of $\alpha_{c}$.
The $\alpha$ values are further normalized such that $\sum_{c=1}^{N} \alpha_{c} = 1$,
where $N$ is the number of available configurations.
Then, the reward function can be formulated as: \looseness=-1 \
\begin{equation}
        r_t(s, e) = \begin{cases}
            \beta - \alpha_{curr} & \text{if } \exists i \in [s,e): \mathbb{L}_A(i)=1\\
            \alpha_{curr}  & \text{if } \forall i \in [s,e): \mathbb{L}_A(i)=0, \\
       \end{cases} \quad
       \label{eqn:local-reward}
\end{equation}

\noindent where $\mathbb{L}_A(i)$ is the label function (\autoref{sec:overview-workflow}), and
$\beta$ is cutoff that divides the configuration space into fast and
slow configurations.


%
The intuition behind this function is that for action frames ($\mathbb{L}_A(i)=1$), the reward should be
inversely proportional to the fastness $\alpha$ since we desire slower more accurate configs.
Conversely, for non-action frames ($\mathbb{L}_A(i)=0$), the reward should be
proportional to $\alpha$.
Thus, the function only looks at the \textit{local} ground truth values to assign
the reward.
With this reward function, the agent checks for the existence of ground truth frames
in the local segment window $(s, e)$.
If there is an action frame in this window (\cref{fig:penalty}), 
the agent penalizes configurations that are fast (low resolution,
coarsely-sampled) and rewards slow configurations.

On the other hand, if there is no action frame in this window, the agent rewards
fast configurations. 
For example, in \cref{fig:lowreward}, the reward assigned to the decision is
positive, but has a low value. 
This is because the agent does not utilize the full window available before the
start of the action. 
In contrast, the decision in  \cref{fig:highreward} receives the maximum reward.
Notice that the agent does not penalize slow configurations when there is no
action in this window.
This design prioritizes the reduction of false negatives
over performance.
This technique does not have explicit control over the query accuracy, similar
to the \heuristic approach.
However, it scales well with multiple knob settings
since it is automated. 

\subsection{Aggregate Reward Allocation}
\label{sec:aggregate-reward-allocation}

\cref{alg:dqn} trains the agent to accelerate the query while penalizing
\textit{any} decision that leads to missing action frames (\cref{fig:penalty}).
So, the agent optimizes for reducing false negatives (instead of performance).
This often leads to a much higher accuracy than the target
at reduced throughput.
So, the \opt can further lower execution time using this
excess accuracy budget. \looseness=-1

\PP{Baseline}
To satisfy the accuracy constraint, \windowpp picks the configuration
that is closest to the target accuracy (\ie just above the required accuracy)
for all the segments.
However, this approach is sub-optimal.
To achieve an aggregate accuracy of 80\% with a better throughput, the agent may
pick configurations with 90\% and 70\% expected accuracy for video segments that
contain and that do not contain actions, respectively.
Since actions are typically infrequent in videos, this strategy leads to higher
throughput by using the faster configuration most of the time, and occasionally
using the slower, more accurate configuration.
Thus, the \opt must take the accuracy constraint into consideration during
the training process to obtain maximum performance.

\PP{Reward aggregation}
The \rl agent must learn the optimal strategy for picking configurations across
the video to minimize query execution time within the given accuracy budget.
To tackle this problem, the \opt uses an accuracy-aware aggregate reward allocation strategy.
Here, the \opt accumulates the agent's decisions during a predetermined window $W$.
During this accumulation phase, the algorithm does not immediately assign a
reward to the agent.
Specifically, in \cref{alg:dqn:gen-exp-reward} of \cref{alg:dqn}, the agent
only receives a reward after it processes all the frames in $W$.
At the end of the window, the aggregate accuracy of the agent's decisions is
evaluated.
Based on the actual and target accuracy metrics, it computes the reward
for all the local decisions taken within this window.
\rev{We use the term \textit{aggregate} to emphasize that we do not immediately assign local rewards as in~\cref{eqn:local-reward}.
We instead accumulate decisions during each window, then compute the reward for this window (shown in~\autoref{sec:accuracy-based-reward}),
and finally assign the (same) reward to all the decisions in the window.}

\begin{algorithm}[t]
\SetArgSty{textnormal}
\footnotesize
\setstretch{1.2}
  \SetKwInOut{Input}{Input}
  \SetKwInOut{Output}{Output}
  \SetKwInOut{Initialize}{Initialize}
  \SetKwProg{myproc}{Procedure}{}{}
  \Input{Video V, Target Accuracy $\alpha$, Window W $\rightarrow (s,e)$\\
         $C \rightarrow \{c_1, c_2, \cdots, c_k\}$\\ 
         Binary frame-wise ground truths GT(W) of size $|e-s|$ \\
         Trained APFG U}
  \Output{List of size \texttt{k} with scalar rewards for each decision in C}
  \Initialize{
        Pred(W) $\gets$ Empty list of size $|e-s|$\;
  }
  \myproc{GetReward(V, $\alpha$, C)}
      { %
        \tcp{Collect predictions}
        \For{i $\gets$ $1$ to k}{   \label{alg:reward:start-collect-pred}
            \texttt{segment} $\gets$ GetSegment(V, c\textsubscript{i})\;
            \texttt{pred(i)} $\gets$ U(\texttt{segment}, c\textsubscript{i})\;
            Update\_Prediction\_List(\texttt{pred(i)})\;
        }\label{alg:reward:end-collect-pred}
        $\alpha\prime \gets$ Accuracy(GT(W), Pred(W))\; \label{alg:reward:compute-window-accuracy}
        \tcp{Assign rewards}
        \uIf{$\alpha' \ge \alpha$}{ \label{alg:reward:start-assign-reward}
            $r_i \gets \frac{1 - \alpha'}{1 - \alpha}   \forall i \in [1,k] $\; \label{alg:reward:pos-reward}
            }
        \uElse(){ 
            $r_i \gets \alpha' - \alpha \forall i \in [1,k]$\; \label{alg:reward:neg-reward}
        }
      }
  \caption{Accuracy-Aware Reward Assignment Algorithm}
  \label{alg:reward}
\end{algorithm}

\subsection{Global Accuracy-Aware Rewards}
\label{sec:accuracy-based-reward}
%

%
As discussed in \autoref{sec:aggregate-reward-allocation}, local rewards are not
sufficient to ensure that \sys efficiently meets the global target accuracy
constraint. 
Hence, we modify the reward function used in~\cref{alg:dqn} (\cref{alg:dqn:gen-exp-reward})
to use the aggregate accuracy metric presented in~\autoref{sec:aggregate-reward-allocation}.
\cref{alg:reward} outlines how the \opt computes the accuracy-aware aggregate
reward at the end of each window W.
Assume that the agent takes $k$ steps in this window, then $\{c_1, c_2, \cdots,
c_k\}$ is the sequence of configurations chose by the agent in the $k$ time
steps. 
With the aggregate window approach, the \opt first collects predictions in the window
(\cref{alg:reward:start-collect-pred} to \cref{alg:reward:end-collect-pred}).
Next, it computes the overall accuracy in the window using ground truth labels
(\cref{alg:reward:compute-window-accuracy}).
The \opt then assigns a reward to all the decisions in the window based on 
the accuracy achieved by the agent.

\PP{Updated reward function}
\cref{alg:reward:start-assign-reward} to
\cref{alg:reward:neg-reward} show the updated accuracy-based reward function.
If the target accuracy $\alpha$ is met within the window, the \opt assigns a reward
that increases when the achieved accuracy is closest to the target accuracy
(\cref{alg:reward:pos-reward}).
The intuition behind this reward is that, as long as the accuracy achieved
is greater than the target accuracy, we want to maximize the throughput.
As we have seen in~\cref{tab:example-configs}, the throughput is maximum when the
accuracy is minimum.
When the gap between achieved and target accuracy is as small as possible,
the achieved accuracy is minimized, as long as it is above the target.
So, the throughput is maximized while respecting the accuracy constraint.
For example, if the accuracy within the window exceeds target accuracy, the \opt
assigns a low reward, ensuring that the agent tries to exploit this excess
accuracy.
Conversely, if the accuracy in the window is less than the target accuracy, the \opt
assigns a penalty (negative reward) that is directly proportional to the
accuracy deficit (\cref{alg:reward:neg-reward}).
So, the lower the accuracy of the agent, the greater the penalty.

To compute aggregate rewards, \opt uses a delayed replay buffer update strategy
to collect experiences.
During the processing of the current aggregation window, \opt uses
\cref{alg:dqn} to collect the incomplete experience tuples (without reward) into
a temporary buffer.
At the end of each window, the agent updates the experience tuples in the
temporary buffer with the rewards collected using~\cref{alg:reward}.
\sys then pushes the updated experience tuples to the replay buffer.
With these modifications to~\cref{alg:dqn}, \sys achieves a higher
throughput than \windowpp, while also meeting
the accuracy target.

\rev{\PP{Accuracy Guarantees}
Since \sys combines different configurations of \rthreed (\autoref{sec:cost-estimation}), the final accuracy achieved is
dependent on the accuracy achieved by the \rthreed model with individual configurations (\cref{tab:example-configs}).
\sys utilizes accurate configurations for action frames and less accurate configurations for non-action frames.
Thus, \sys ensures a high recall in the output at the cost
of slightly lower precision.
For a given accuracy target, the reward function in \sys balances the precision and recall 
such that the resultant F1-score is closest to the target.
On the other hand, \windowpp is the only other baseline that reaches the accuracy target, since we manually select such configuration from the configuration planning phase~\autoref{sec:cost-estimation}.
\heuristic cannot reliably reach the accuracy target since the rules do not provide explicit control over the accuracy.
}

\rev{
Like state-of-the-art baselines~\cite{kang13blazeit,kang2017noscope,lu2018accelerating}, approximate ML inference queries cannot guarantee accuracy on unseen test data.
So, all the baselines (including \windowpp) cannot be guaranteed to reach the accuracy target on unseen data.
However, we empirically show that \sys meets the target accuracy (\eg target=75-85\% in~\autoref{sec:eval-endtoend} and~\autoref{sec:eval-accuracy-target}) when test and train
data distributions are similar, while other baselines often fail to do so.
We further show that \sys maintains its throughput gains over other methods when trained on one dataset and tested on a different dataset~(\autoref{sec:eval-dom-adapt}) with
 a slight accuracy drop in all the methods (including \windowpp) due to the variations in data distribution.
%
}

%% file: implementation.tex
\section{Implementation}
\label{sec:implementation}


%
%
%

\PP{\apfg Training}
\label{sec:implmentation::udf}
Recall from~\autoref{sec:execution} that the objective of the \apfg is:
(1) to predict the action class labels, and 
(2) to generate features for a given input segment.
%
%
%
\sys trains the \apfg using the segments extracted from a set of training videos
along with their binary labels (\ie whether a particular action class is
present or absent in the segment).
It adopts a fine-tuning strategy to train the \apfg to accelerate the training
process.
Specifically, it uses the pre-trained weights of the \rthreed model trained on the
Kinetics-400 dataset~\cite{carreira2017quo} and fine-tunes the parameters 
for the target action classes.
We add three fully-connected layers to the original \rthreed model.
%
%
%

\PP{Model reuse}
Recall that the \apfg is a collection of action recognition models
that can processing varying resolutions and segment lengths.
We optimize the \apfg training process by approximating the ensemble of
models with a single \rthreed model.
That is, we use one \rthreed model with the highest accuracy to generate
\feature{s} for all configurations.
%
The intuition behind this design is that the model trained on the most accurate
configuration (\ie with the highest resolution and lowest sampling rate) can also
process faster configurations (\eg lower resolutions), albeit at a slight accuracy
drop.
Note that the alternative, \ie re-training the model on lower resolution
configurations would also incur an accuracy drop due to the low resolution,
and would significantly increase the training time.
Although the empirically observed accuracy for the latter strategy is slightly higher,
the former approach leads to much lower training times.
Since the \apfg allows model flexibility, assuming that the computational resources
are available, one could also train the models individually
for the configurations.

\PP{\rl Training}
\label{sec:implmentation::rl}
To train the \rl agent, \sys use the DQN algorithm with an experience replay
buffer.
We configure the replay buffer to 10~K experience tuples and initialize it with
5~K tuples.
\sys{'s} DQN model is a Multi-layer Perceptron (MLP) with 3 fully-connected
layers. 
We use a batch size of 1~K tuples to train the DQN model.
%
%
%
We concatenate all the individual training videos to create episodes of equal time span.
This ensures that the agent receives similar rewards during each episode.
This design decision is critical to optimize the behavior of the \rl agent over 
time.
Additionally, \sys permutes the videos in a random order for each episode
to prevent overfitting to the specific order in which the \rl agent processes the videos. \looseness=-1
%

%

%

\PP{Pre-Processing}
\rl training typically takes more time to converge compared to supervised deep
learning algorithms.
This is because of the large number of experience tuples required for training
the Q-network.
To accelerate this training process, \sys first runs the \apfg on all the input
segments at different resolutions and segment lengths to generate the feature
vectors.
This \textit{preprocessing step} uses a batching optimization and leverages
multiple GPUs to lower the \rl training time significantly.
%
%
The agent then directly uses the precomputed features during training.

\PP{Software Packages}
\label{sec:implmentation::libraries}
We implement \sys in Python 3.8.
We use the PyTorch deep learning library (v 1.6) to train and execute the 
neural networks.
\sys uses the \textsc{R3D-18} model with pre-trained weights from the
PyTorch's TorchVision library~\cite{torchvision},
and fine-tunes it to the evaluation datasets (\autoref{sec:eval-setup}).
%
%
The weights of the \apfg are frozen during the subsequent \rl training
process.
We use the OpenAI Gym (v 0.10.8) library for simulating the \rl
environment~\cite{brockman2016openai}.

%% file: evaluation.tex
\section{Evaluation}
\label{sec:evaluation}


In our evaluation, we illustrate that:
\squishitemize
    \item \sys executes queries up to 4.7$\times$ faster than the state-of-the-art
    baseline (\windowpp) at a given target accuracy (\autoref{sec:eval-endtoend}),  8.3$\times$ higher throughput and $0.25$ points higher F1 score on average compared to existing VDBMSs (\framepp) (\autoref{sec:eval-endtoend}),
    with a maximum speedup of 22.1$\times$.
    \item \sys consistently achieves the user-specified accuracy target at a better
    throughput than other baselines (\autoref{sec:eval-accuracy-target}).
    Meanwhile, \sys's knobs contribute to the throughput increase (\autoref{sec:eval-knob-selection}).
    \rev{\item The RL agent in \sys is practical -- a single model can be trained to detect multiple similar actions (\autoref{sec:eval-practicality})
    and the same model can detect a given action in multiple datasets (\autoref{sec:eval-dom-adapt}).}
    \item \sys's \opt incurs only a small overhead for training the RL agent (\autoref{sec:eval-training-costs}) and leveraging available configurations for performance improvement
    (\autoref{sec:eval-cfg-dist}).
\squishend
%

\begin{table}[t!]
    \small
    \centering
    \begin{adjustbox}{width=1.05\columnwidth,center}
    \begin{tabular}{cccccccc}
    \toprule
    \textbf{Dataset} & \textbf{No. of}  & \textbf{Total} & \textbf{Percent} &
    \textbf{Avg. Action} & \textbf{Std.} & \textbf{(Min, Max)}\\
    &  \textbf{classes}  & \textbf{ Frames (K)} & \textbf{Actions} &
    \textbf{length} & \textbf{Dev.}& \textbf{Length}\\ \midrule
    BDD100K & 2 & \num[group-separator={,}]{186} & 7.03 & 115 & 58.7 & (6, 305)
    \\
    Thumos14 & 2 & \num[group-separator={,}]{645}  & 40.27 & 211 & 186.3 & (18, 3543)\\
    ActivityNet & 2 & \num[group-separator={,}]{633} & 56.37 & 909 & 1239.1 &
    (20, 6931) \\\bottomrule
    \end{tabular}
    \end{adjustbox}
    \caption{
        \textbf{Dataset Characteristics} -- 
        We report these key characteristics of the datasets we evaluate \sys on:
        (1) number of action classes and total frames,
        (2) average percentage of action frames across all frames,
        (3) average and standard deviation of length of action instances, and
        (4) length of shortest and longest action instances.
        }
    \label{tab:dataset-stats}
\end{table}

\subsection{Experimental setup}
\label{sec:eval-setup}

\PP{Evaluation queries and dataset}
We evaluate \sys on six queries from three different action localization
datasets.
For the first four queries, we use publicly available datasets:
Thumos14~\cite{THUMOS14} and ActivityNet~\cite{caba2015activitynet}.
Each has hundreds of long, untrimmed videos collected from several
sources.
We take two classes from each dataset to construct four action queries -
\squishitemize
\item Thumos14 - \polevault, \cleanandjerk
\item ActivityNet - \ironingclothes, \tennisserve
\squishend

However, these datasets contain high density of actions (\ie more than 40\%
of the video frames contain actions) (see~\autoref{tab:dataset-stats}). 
To better study the impact of action length and percentage of action frames
(\abbrev action percentage), we leverage a novel dataset tailored for evaluating
action analytics algorithms.
This dataset consists of a subset of 200 videos from the BDD100K (Berkeley
Deep Drive) dataset~\cite{yu2020bdd100k}.
Each video is $\approx$ 40 seconds long, collected at 30~fps, and
contains dash-cam footage from cars driving in urban locations.
We manually annotate these videos with two action classes/queries:
\squishitemize
\item \crossright: In this query, the pedestrian crosses the street
from the left side of the road to the right side (\cref{fig:example}).
\item \leftturn: This query returns the segments in the video that contain
a left turn action the driver's point-of-view.
\squishend

\PP{Dataset Characteristics}
\autoref{tab:dataset-stats} lists the key characteristics of the datasets.
The datasets vary significantly in action percentage and average
action length.
ActivityNet has the highest action percentage, and the action length varies
significantly across videos (high std. deviation).
BDD100K has shorter actions and lower action percentage
(only 7\% of all the frames are action frames). \looseness=-1

\PP{Configurations}
We list the available knob settings for each dataset
in~\autoref{tab:config-stats}.
Each unique combination of knob settings represents a candidate configuration.
BDD100K dataset has a total of 64 available configurations
(4$\times$4$\times$4), while Thumos14 and ActivityNet each has 27.
Notice that the available segment lengths (window size)
are longer for queries with higher action length and percentage.
This results in a higher overall throughput for the datasets/queries with
large action lengths/percentages (\autoref{sec:eval-endtoend}).
On the contrary, we pick shorter segment lengths for BDD100K,
since the action length and percentage is significantly smaller
(\autoref{tab:dataset-stats}).
When the action length is shorter, the accuracy drops at larger window sizes.
For example, if we use the configurations for ActivityNet (in table 4)
on BDD100K, the accuracy drops rapidly because the configurations with large
windows just completely skip the action.

\autoref{tab:config-stats} also lists the highest accuracy achieved by
\textit{any} configuration for each query.
This denotes the upper bound on accuracy achievable by any of the query
processing techniques for the query.
The best accuracy is higher for BDD100K compared to Thumos14 and ActivityNet
datasets.
This is likely because: 
(1) BDD100K has relatively simple classes (\eg left turn vs tennis serve),
(2) It has higher resolution videos (300x300 vs 160x160), and
(3) videos are collected from the same camera and environment.

\begin{table}[t!]
    \centering
    \begin{adjustbox}{width=1.05\columnwidth,center}
    \begin{tabular}{cccc|cc}
    \toprule
    \textbf{Dataset} & \textbf{Available}  & \textbf{Available} & \textbf{Available} &
    \multicolumn{2}{c}{\textbf{Maximum}}\\
    &  \textbf{Resolutions}  & \textbf{ Segment Lengths} & \textbf{Sampling Rates} &
    \multicolumn{2}{c}{\textbf{Accuracy}}\\ \midrule
    \multirow{2}*{BDD100K} & \multirow{2}*{150, 200, 250, 300} & \multirow{2}*{2, 4, 6, 8} & \multirow{2}*{1, 2, 4, 8} & Cross Right & 0.91\\
    & & & & Left Turn & 0.89\\
    \midrule
    \multirow{2}*{Thumos14} & \multirow{2}*{40, 80, 160} & \multirow{2}*{32, 48, 64} & \multirow{2}*{2, 4, 8}  & Pole Vault & 0.78\\
    & & & & Clean And Jerk & 0.76\\
    \midrule
    \multirow{2}*{ActivityNet} & \multirow{2}*{40, 80, 160} & \multirow{2}*{32, 48, 64} & \multirow{2}*{2, 4, 8}  & Ironing clothes & 0.85\\
    &  &  &  & Tennis serve & 0.80\\\bottomrule
    \end{tabular}
    \end{adjustbox}
    \caption{
        \textbf{Configuration Statistics} --
        We report these statistics for each dataset we evaluate \sys on:
        (1) Number of available resolutions to choose from,
        (2) Number of available segment lengths,
        (3) Number of available sampling rates, and
        (4) Maximum accuracy achieved by \textit{any} configuration for a given
        query.
        }
    \label{tab:config-stats}
\end{table}

\PP{Hardware Setup}
We conduct the experiments on a server with one NVIDIA GeForce RTX 2080
Ti GPU and 16 Intel CPU cores. 
The system is equipped with 384 GB of RAM. 
We use four GPUs for training and extracting features. 
We constrain the number of GPUs used during inference to one to ensure a fair
comparison.

\PP{Baselines}
We compare \sys against three baselines.
\squishitemize
\item \framepp uses a 2D-CNN on individual frames in the video (\autoref{fig:framepp})
and outputs a binary label that determines the presence or absence of an action
in that frame. 
Existing VDBMSs~\cite{kang2017noscope, lu2018accelerating, kang13blazeit} use
\framepp as a filtering optimization to accelerate object queries. 
To improve accuracy on action queries, we instead apply \framepp on all frames.
\item \rev{\segmentpp uses a lightweight 3D-CNN filter on all non-overlapping segments in the video
to quickly eliminate segments that do not satisfy the query predicate.
The \rthreed model then processes the filtered segments to generate the final query output.
\segmentpp extends the frame-level filtering optimization used in existing VDBMSs~\cite{kang2017noscope, lu2018accelerating, kang13blazeit} to segments.
}
\item \windowpp processes segments in the video using a
3D-CNN, specifically, the \rthreed network~\cite{6165309, tran2018closer}.
We use the network in a sliding window fashion (\autoref{fig:sliding})
on the input video to generate segment-level predictions.
\windowpp uses a static \configuration for the entire dataset.
It chooses the fastest configuration that meets the target accuracy.
\item \heuristic dynamically uses a subset of available configurations
based on hard-coded rules to process the video, including
(1) using the slowest configuration when the \apfg returns \texttt{ACTION}
prediction, 
(2) a faster configuration when the \apfg prediction
flips from \texttt{ACTION} to \texttt{NO-ACTION}, and
(3) the fastest configuration when the \apfg returns a
\texttt{NO-ACTION} prediction across ten consecutive time steps.
\item We refer to the \rl-based approach as \sysrl.
\squishend

\begin{figure}[t!]
    \centering
    \hspace{2em}\includegraphics[width=0.9\linewidth]{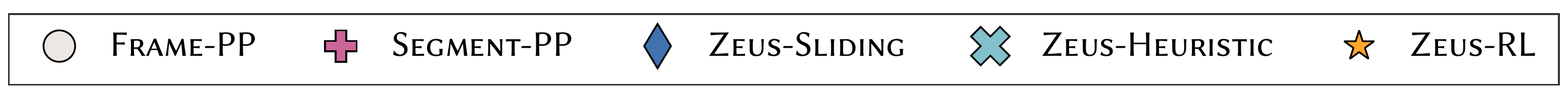}
    \includegraphics[width=0.85\linewidth]{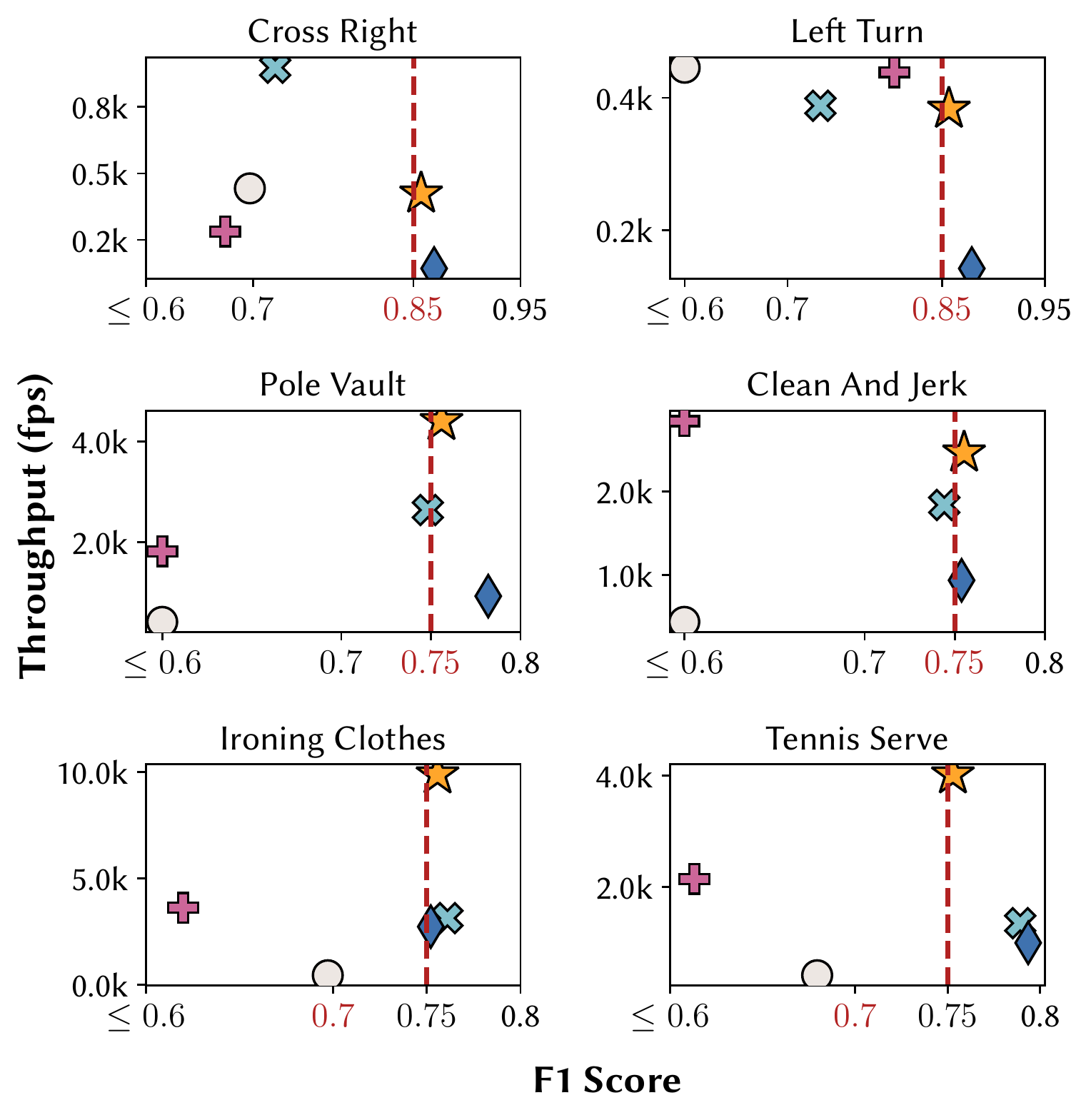}
    \caption{
     %
     Throughput and F1 score achieved by \sysrl, \framepp, \rev{\segmentpp}, \heuristic, and
     \windowpp over \qa-\qf (Top-right corner is best). 
     Note that ranges are different for each query. 
     The red dotted line represents the accuracy target set for each query.
     }
    \label{fig:endtoend}
\end{figure}

\subsection{End-to-end Comparison}
\label{sec:eval-endtoend}

%
We first  compare the end-to-end throughput and accuracy metrics of \sysrl
against the other baselines\footnote{
We do not report the pre-processing time for all the techniques since it is
parallelizable on the CPUs.
}.
We use accuracy target of $0.75$ for queries based on Thumos14 and
ActivityNet, while $0.85$ for queries based on BDD100K
(due to the maximum accuracy that can be achieved~\autoref{tab:config-stats}). 
In this experiment, \framepp uses the most accurate trained model (and hence
with the highest available resolution).
\windowpp uses the fastest configuration that achieves the accuracy target
on training data.
\heuristic applies hard-coded heuristics on a subset of configurations
that are used by \sysrl to process the query.

\PP{\windowpp}
The results are shown in~\cref{fig:endtoend}.
The red dotted line indicates the accuracy target for each query.
Notably, on average, \sysrl is $3.4\times$ faster
than \windowpp at a given target accuracy, with a maximum speed-up of
$4.7\times$ in \polevault.
%
\sysrl outperforms \windowpp across all queries at the target accuracy.
This shows that the additional configurations available to \sysrl help in
improving performance.
\windowpp is the only other baseline that reaches the accuracy target
for all the queries.
However, the accuracy achieved is often higher than the target (\eg Tennis serve).
In most cases, \sysrl achieves accuracy closest to the target accuracy.
It efficiently uses the excess accuracy to improve throughput by using
faster configurations, thus outperforming \windowpp.

\PP{\heuristic}
For four out of the six queries, \sysrl has a higher throughput than \heuristic.
The throughput delivered by \heuristic is inversely proportional to the
percentage of actions in the input videos.
Consider the \crossright (low percentage) and \tennisserve (high percentage)
queries. 
\heuristic returns a low throughput in \tennisserve and a high throughput in
\crossright (albeit at a lower accuracy).
The reason for this behavior is that the hard-coded rules in \heuristic
state that the agent must pick the slower configurations for \texttt{ACTION}
frames.
So, \heuristic does not have explicit control over query accuracy.
When the fraction of action frames in the input video is high (\eg
\tennisserve), \heuristic uses slower configurations for majority of the frames
in the video.
So, it delivers a lower throughput while overshooting the accuracy target.  

On the contrary, \sysrl does not have hard-coded rules. The \rl agent is trained
to automatically pick configurations that improve throughput while
\textit{barely exceeding} the accuracy target.
This also explains why \heuristic has a high throughput for \crossright.
This query contains fewer, shorter action segments, resulting in \heuristic
choosing the faster configurations most of the time, thereby
skipping important frames (due to large window sizes).
We show the concrete distribution of these configurations in~\autoref{sec:eval-cfg-dist}.
Finally, \heuristic performs better than \windowpp across most queries.
This illustrates that dynamically selecting the configuration is better than
using a static configuration.
To summarize, (1) an RL-based agent is more efficient at selecting the optimal
configuration at each time step than a rule-based agent, 
(2) dynamic configuration selection is better than a static configuration.

\rev{
\PP{\segmentpp}
\segmentpp returns a prohibitively low F1 score ($\le 0.6$) for four out of the six queries.
It returns a low accuracy for the harder action classes from the Thumos14 and ActivityNet datasets.
This behavior can be attributed to the inability of the lightweight filters in \segmentpp to capture the inherent complexity of actions.
On the other hand, notice that \segmentpp provides a better accuracy on the easier \leftturn class.
The throughput achieved by \segmentpp is a function of the action percentage in the videos.
When action class is trivial and rare (\eg \leftturn), \segmentpp can rapidly eliminate segments that do not satisfy the predicate, thus providing a slightly better throughput.
However, when the action classes are hard, the lightweight filters in \segmentpp are highly inaccurate (with F1 score as low as 0.2).
In case of hard and rare classes (\eg \crossright), this results in increased false negatives and false positives, thus leading to a much lower F1 score and slightly lower throughput.
Finally, when the action classes are hard and frequent (\eg \ironingclothes), the increased false negatives and false positives lead to both, a much lower throughput and accuracy than \sys.
}

\PP{\framepp}
\sysrl is $8.3\times$ faster and delivers $0.25$ points
higher F1 score on average compared to \framepp.
\sysrl uses segment-level processing to process up to 64 frames with a single \apfg
invocation.
On the other hand, \framepp has to process each frame separately.
So, even though each \apfg invocation is 5.9$\times$ faster in \framepp (\autoref{sec:action-localization}),
the overall throughput of \sysrl is significantly higher.
Inspite of processing each individual frame, \framepp still has a
prohibitively low F1 score for all the queries.
This is because these queries require a sequence of frames (temporal
information) to predict the action class, and hence cannot be handled by
\framepp.

\PP{Impact of Action Length and Action Percentage}
The relative throughput of all the methods varies across datasets.
We attribute this to variability in action length and action percentage across
datasets (\autoref{tab:dataset-stats}).
The average speedup of \sysrl over \windowpp is the highest for ActivityNet
(3.8$\times$) and lowest for BDD100K (2.8$\times$). 
So, \sysrl performs better when the action length and percentage are higher.
For dataset with longer action sequences, we set a larger segment length and
sampling rate (\autoref{tab:config-stats}) for more efficient processing. 
So, each \apfg invocation results in larger processed windows, resulting in
proportionally larger absolute and relative throughput.

\begin{figure}[t!]
    \centering
    \includegraphics[width=0.45\linewidth]{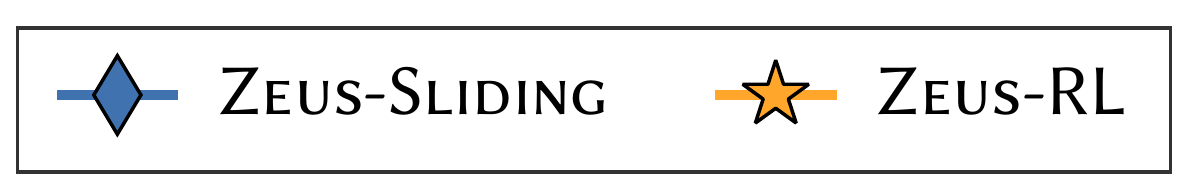}
    \includegraphics[width=0.85\linewidth]{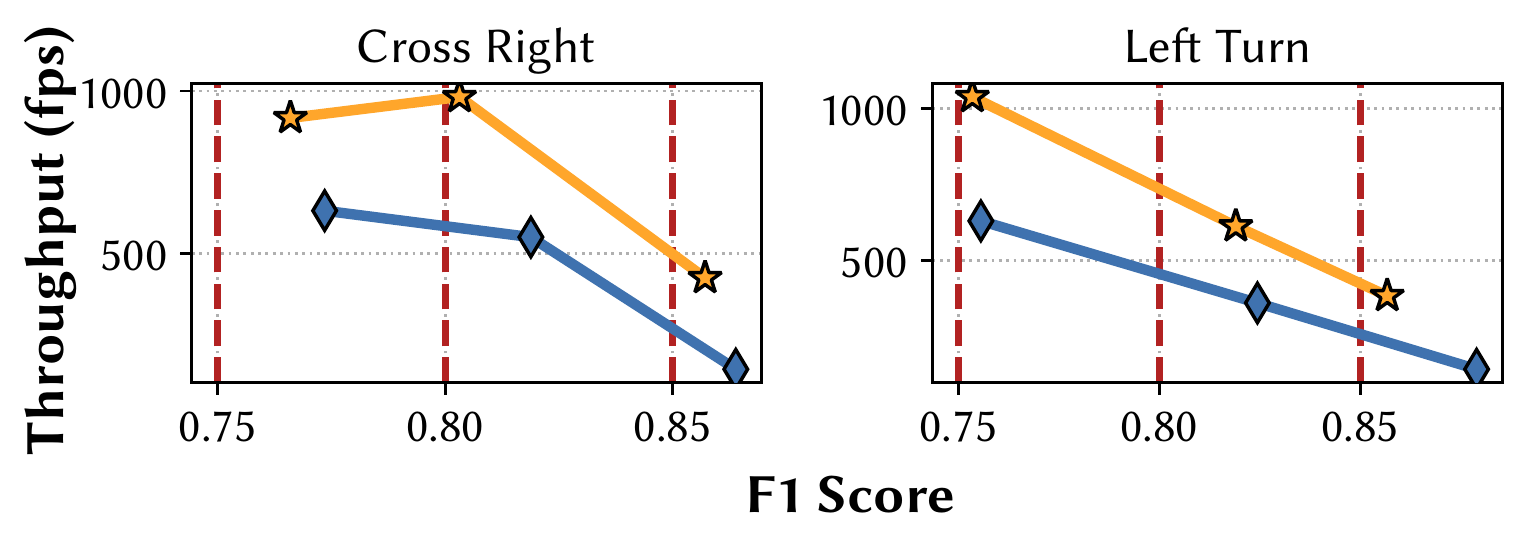}
    \caption{
     %
     Performance and throughput of \windowpp and \sysrl over a range of target accuracies.}
    \label{fig:accuracy-target}
\end{figure}

\begin{table}[t!]
    \small
    \centering
    \begin{adjustbox}{width=1.05\columnwidth,center}
    \begin{tabular}{lll|lll}
    \toprule
    \multicolumn{3}{c|}\crossright & \multicolumn{3}{c}\leftturn \\\midrule
    \textbf{Accuracy} & \textbf{Accuracy} & \textbf{Speedup} & \textbf{Accuracy} & \textbf{Accuracy} & \textbf{Speedup}\\
    \textbf{Target} & \textbf{Achieved} & & \textbf{Target} & \textbf{Achieved} & \\ \midrule
    0.75 & 0.753 & 1.45 \tikzmark{a} & 0.75 & 0.755 & 1.64 \tikzmark{c} \\
    0.80 & 0.819 & 1.78 & 0.80  & 0.824 & 1.70\\
    0.85 & 0.857 & 2.97 \tikzmark{b} & 0.85 & 0.879 & 2.69 \tikzmark{d} \\\bottomrule
    \end{tabular}

    \tikz[remember picture,overlay] \draw[->,thick] (a.center -| b.center) -- (b.center);
    \tikz[remember picture,overlay] \draw[->,thick] (c.center -| d.center) -- (d.center);
    \end{adjustbox}
    \caption{
        \textbf{Speedup of \sysrl over \windowpp over a range of accuracy targets}
        }
    \label{tab:aggregate-speedup}
\end{table}

\subsection{Accuracy-Aware Query Planning}
\label{sec:eval-accuracy-target}

In this experiment, we verify the ability of the \opt to generate an
accuracy-aware query plan.
\cref{fig:accuracy-target} shows the throughput vs accuracy curve for \sysrl and
\windowpp at three accuracy targets over two queries.
The red dotted line indicates the three different accuracy targets for
the queries (0.75, 0.80, 0.85).
\sysrl performs better than \windowpp over all the accuracy targets.
Furthermore, \sysrl accuracy is consistently closer to the target accuracy,
while \windowpp often overshoots this target.
\sysrl efficiently allocates this excess accuracy budget by using faster
configurations in certain segments to improve throughput, while
ensuring that the accuracy is just over the target.
~\autoref{tab:aggregate-speedup} shows the speedup of \sysrl over \windowpp
at each accuracy target.
The speedup is inversely proportional to the accuracy target.
Recall that \sysrl uses the \feature{s} generated by the \apfg to
select the next configuration.
When the accuracy target is low, the configurations available to \sysrl
are all of low accuracy.
So, \sysrl receives noisy features from the \apfg, resulting in sub-optimal
configuration selection.

\subsection{Knob Selection}
\label{sec:eval-knob-selection}

\begin{figure}[t!]
    \centering
    \includegraphics[width=0.8\columnwidth]{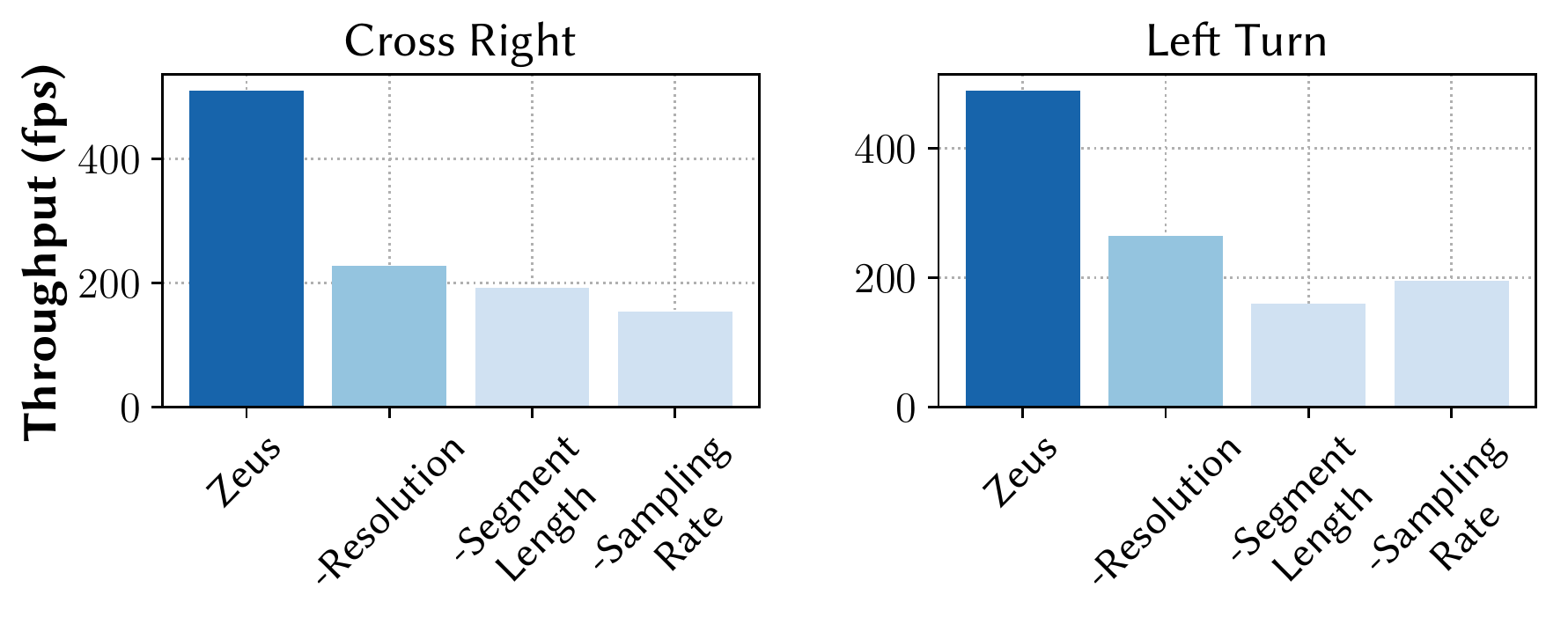}
    \caption{
     %
     Impact of \resolution, \segmentlength, and \samplingrate knobs on the throughput of \sysrl. We disable
     each knob to study its impact on the overall throughput.}
    \label{fig:ablation}
\end{figure}

We investigate the contribution of different optimizations
(\ie knobs) to the throughput of \sysrl.
Specifically, we disable each knob (fix the value) one at a time and examine its impact on
throughput.
The results are shown in~\cref{fig:ablation}.
Notably, all of the knobs contribute to the performance improvement,
and \segmentlength, \samplingrate are the key knobs.

Across all queries, on average, disabling the \samplingrate, the \segmentlength,
and the \resolution knobs reduces throughput by 62\%, 51\%, and 36\%
respectively.
Note that the \samplingrate and \segmentlength knobs operate in tandem to
determine the throughput of \sysrl.
For example, when \segmentlength is 8 and \samplingrate is 8, the agent
processes 64 frames in one time step.
In contrast, when \segmentlength is 2 and \samplingrate is 1, it only processes
2 frames in one time step.

\PP{Importance of Knobs}
We now examine why the \samplingrate and the \segmentlength knobs are important.
The throughput of \sysrl is determined by two factors:
(1) number of \apfg invocations, and
(2) time taken for each \apfg invocation.
Among these factors, the first one is more important.
When \sysrl processes 10 frames using one \apfg invocation (\ie \segmentlength =
10), it is 2$\times$ faster than using 5 \apfg invocations (\ie \segmentlength =
2). 
The \samplingrate and the \segmentlength knobs reduce the number of \apfg
invocations, thereby improving throughput.
The \segmentlength knob determines the size of the input along the
temporal dimension.
A longer segment leads to higher \apfg invocation time.
So, the impact of this knob is comparatively smaller than that of the
\samplingrate knob.
The impact of the \resolution knob is the least since it only affects the \apfg
invocation time.

\PP{\apfg Invocation Time}
We attribute the lower impact of each \apfg invocation time to the lack of input
batching.
Since the \rl agent is able to generate the next input only after processing the
current input, it cannot exploit the batching capabilities of the GPU.
So, \sysrl does not support intra-video parallelism.
But, it is possible to extend \sysrl to support inter-video parallelism.
Here, batching inputs across videos would allow better GPU  utilization.

\subsection{\rev{Practicality of \sys}}
\label{sec:eval-practicality}
\begin{figure}[t!]
    \centering
    \includegraphics[width=0.85\linewidth]{figures/endtoendlegendrevision.pdf}
    \includegraphics[width=0.8\linewidth]{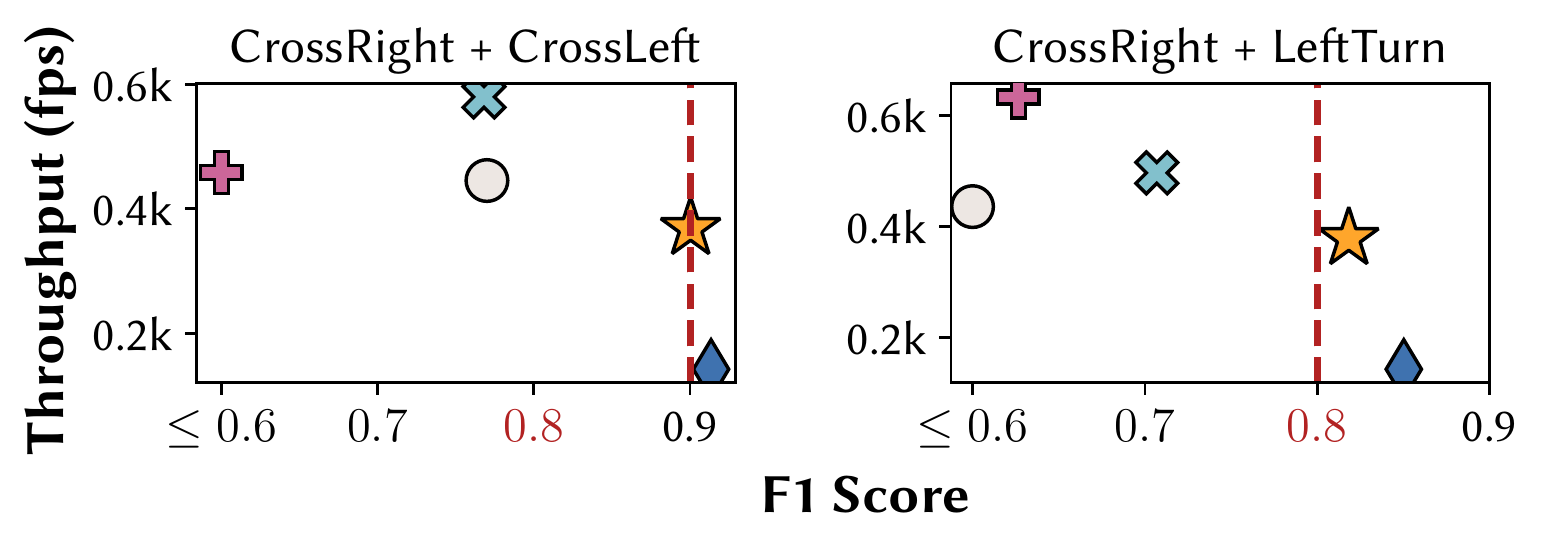}
    \caption{
     \rev{\textbf{Multi-Class Training} -- Throughput and F1 curve for different
     methods trained on multiple classes together.}}
    \label{fig:multi-class-train}
\end{figure}

\rev{In this section, we study the practical applicability of \sys. \\
\PP{Multi-class training}
Since training one single RL agent for each action class may not be viable, we first examine the ability of \sys to train on multiple classes together.
More precisely, we combine the ground truths of the two action classes such that frames belonging to either of the action class are considered true positives and frames that belong to neither are considered true negative.
Once the models return the output segments for both the classes, we can trivially separate them using another classifier since the output segments are only 3-4\% of the full videos.
We show the results in~\cref{fig:multi-class-train}.
Most notably, \sysrl provides the best accuracy-performance trade-off even when trained on two classes.
The performance of all the methods, specifically accuracy, is slightly better for the combination of (\crossright, \crossleft) compared to (\crossright, \leftturn).
In the first case, the action instances are similar looking, thus lowering the task complexity.
In the second case, the actions \crossright and \leftturn are characteristically different, which reduces the accuracy of the \apfg and thus \sysrl.
This also explains the high accuracy achieved by \framepp for the (\crossright, \crossleft) combination.
When these classes are combined, the goal of \framepp reduces to detecting frames that contain a person
in front of the car, making it a simpler task than detecting the walking direction of the person.
}

\begin{figure}
    \centering
    \captionsetup[subfigure]{width=\linewidth,justification=raggedright}
\begin{subfigure}[b]{0.40\columnwidth}
    \centering
    \includegraphics[width=1.2\linewidth]{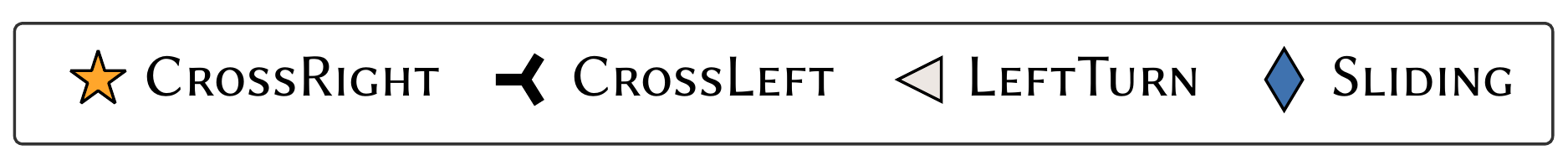}
    \includegraphics[width=\linewidth]{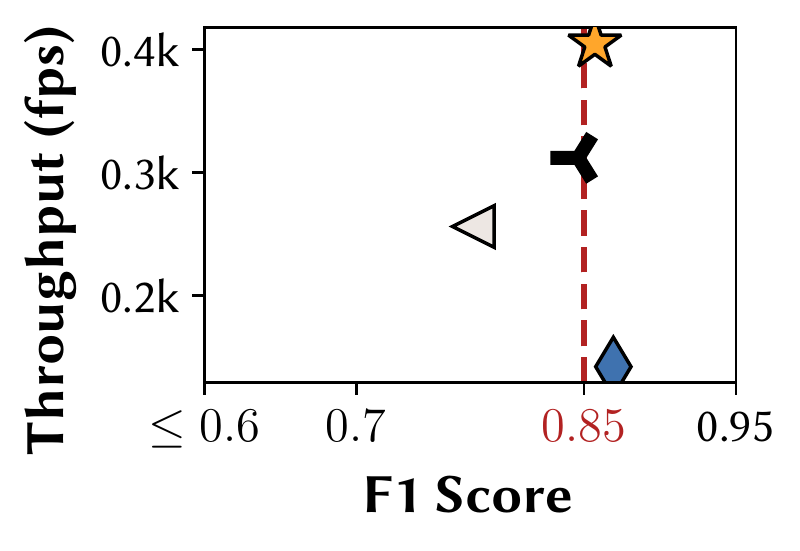}
    \caption{
    \rev{Throughput and F1 score of model trained \\ on \crossright}}
    \label{fig:across-classes-cross-right-to-cross-left}
\end{subfigure}
\hfill
\begin{subfigure}[b]{0.59\columnwidth}
    \centering
    \includegraphics[width=\linewidth]{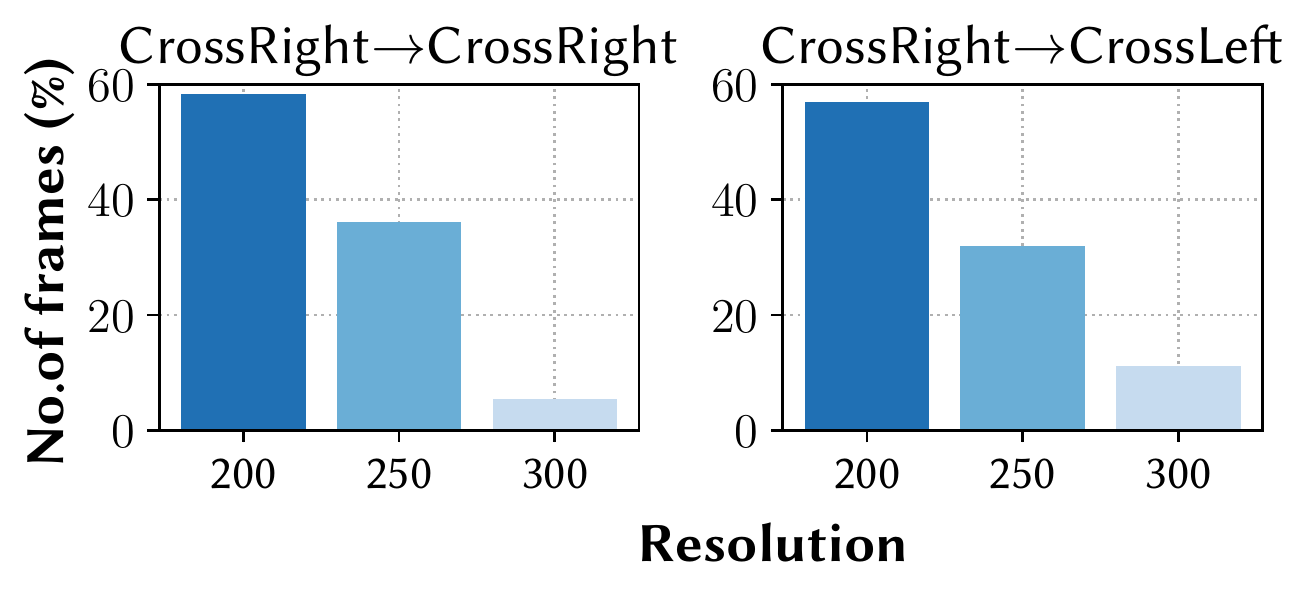}
    \caption{
     \rev{Number of frames processed at each resolution when using the same model for both classes}}
    \label{fig:across-classes-configs}
\end{subfigure}
\caption{
    \rev{\textbf{Cross-Model Inference} -- Evaluation of model trained on
    \crossright over other classes}}
 \label{fig:across-classes}
\end{figure}

\rev{
\PP{Cross-model inference}
In the second experiment, we use the RL agent trained on one class directly for other classes.
More precisely, the RL agent takes decisions based on the features received from
the \apfg models.
Our intuition is that the feature vectors generated by the \apfg for different action classes
are similar, especially for similar looking classes such as \crossright and \crossleft.
So, we can directly use the RL agent trained for one action class on the other action class, just by
using the \apfg models corresponding to each class.
To this end, we train an RL agent on the \crossright class and directly use it for the \crossleft and \leftturn classes.
We show the results in~\cref{fig:across-classes-cross-right-to-cross-left}.
We can see that the model trained on \crossright provides a decent throughput at minimal accuracy loss on the \crossleft query.
In fact, the model achieves a~2.2$\times$ speedup over the \windowpp baseline for the \crossleft query.
The throughput and accuracy are slightly lower when the same model is used for the \leftturn class, since the differences in the action instances
lead to more diverging feature vectors than those that the model is trained on.
Finally, we show the number of frames processed at each resolution in~\cref{fig:across-classes-configs}.
When the \crossright model is used directly for \crossleft, we see that the number of frames processed at the high resolution of 300x300 increases,
while those processed at 200x200 decreases.
The slightly different feature vectors lead to some suboptimal decisions by the RL agent, but the overall trend still remains the same.
}

\subsection{\rev{Domain Adaptation}}
\label{sec:eval-dom-adapt}
\begin{figure}[t!]
    \centering
    \hspace{2em}\includegraphics[width=0.85\linewidth]{figures/endtoendlegendrevision.pdf}
    \includegraphics[width=\linewidth]{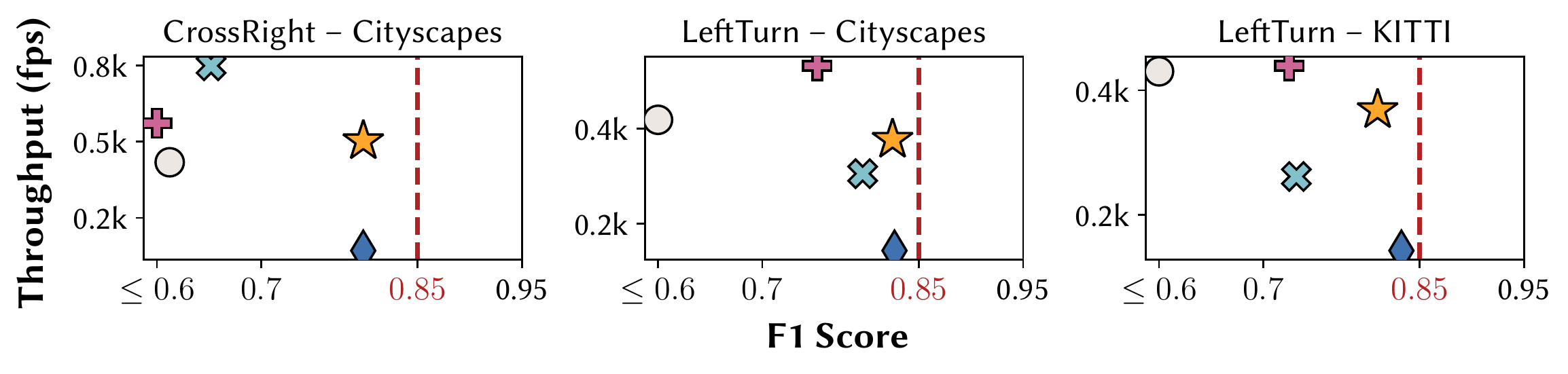}
    \caption{
     %
     \rev{Domain Adaptation -- Throughput and F1 score of \sysrl and other baselines trained on the BDD100k dataset and evaluated on the Cityscapes and KITTI datasets.}}
    \label{fig:eval-domain-adapt}
\end{figure}

\rev{
In this experiment, we evaluate the domain adaptation ability of \sysrl.
Specifically, we train \sysrl and the four baselines on the BDD100K driving
dataset and run the trained models on the Cityscapes~\cite{Cordts2016Cityscapes} 
and the KITTI~\cite{Geiger2013IJRR} datasets. 
BDD100k dataset contains driving scenes from the streets of 4 US cities, while
the Cityscapes dataset contains driving scenes from the city streets of
Frankfurt, Germany. 
The KITTI dataset contains driving scenes from the residential streets of
Karlsruhe, Germany. 
So, the datasets are inherently different in terms of scene composition and
action distribution. 
We use the same experimental setup as~\autoref{sec:eval-endtoend} with an
accuracy target of $0.85$ for both queries. 
We evaluate the \crossright query only on Cityscapes due to no available action instances for this class in the KITTI dataset.
The results are shown in \cref{fig:eval-domain-adapt}.
}

\rev{
Most notably, \sysrl maintains its advantages over the other baselines even when tested on different datasets.
The relative performance of the different baselines remains consistent on the other datasets.
All the methods fail to reach the accuracy target by a small margin ($\sim$2.5\%).
The slight accuracy drop is reasonable considering the challenges of tackling
data drift~\cite{suprem2020odin}.
The accuracy drop is more considerable in \crossright than \leftturn since the former is a more complex action.
Similarly, the accuracy drop for \leftturn is more significant in the KITTI dataset compared to the Cityscapes dataset
since the residential scenes in KITTI lead to more variations in driving and action patterns.
\sysrl provides accuracy on par with the most accurate approach \ie \windowpp and does so at a better performance.
\heuristic suffers a significant accuracy drop in \crossright and a throughput drop in \leftturn.
The inability of \heuristic to balance accuracy and throughput is worsened on the other datasets (compared to ~\cref{fig:endtoend})
because individual decisions of the \apfg are slightly less accurate, leading to higher noise in the efficacy of the rules.
\framepp and \segmentpp achieve the lowest accuracy (as low as 0.15) when tested on the other datasets.
Their inability to capture the complexities of actions is exacerbated when tested on unseen datasets.
}

\subsection{Training Cost}
\label{sec:eval-training-costs}
We now examine the training overhead for \sysrl.
We show the cost of training different components of the pipeline in \autoref{tab:training-costs}.
The \apfg Training cost shown is the cost to train the 2D/3D-CNNs.
Recall that we train the \apfg only once with the best configuration (\ie the highest resolution
lowest sampling rate configuration) (\autoref{sec:implmentation::udf}).
%
%
In the cost estimation phase (\autoref{sec:cost-estimation}), we run the \apfg (using \windowpp technique) with
each configuration on a tiny held out validation set to get the throughput and accuracy of each configuration.
Since this preprocessing cost is \textit{same} for all the techniques, we
omit it from the training cost.

The overall training time of \sysrl is 35\% (90 sec) more than \windowpp,
for training the \rl agent.
This is a one-time preprocessing cost for a given query and accuracy target.
Further, the total training and inference time for \windowpp is still 27\%
higher than \sysrl owing to the faster inference with \sysrl.
Finally, even though the training time of \heuristic is comparable to \windowpp,
\heuristic needs manual construction of rules involving several configurations.
This limits the practical applicability of \heuristic.

\begin{table}[t!]
    \centering
    \begin{adjustbox}{width=\columnwidth,center}
    \begin{tabular}{ccccc}
    \toprule
    \textbf{Method} & \textbf{\apfg Training (s)} & \textbf{RL Training (s)} & \textbf{Inference (s)} \\\midrule
    \framepp & 101.81 & NA & 396.85 \\
    \windowpp & 247.57 & NA &  181.06 \\
    \heuristic & 247.57 & NA & 64.21 \\
    \sysrl  & 247.57 & 90.00 & 38.52 \\ \bottomrule
    \end{tabular}
    \end{adjustbox}
    \caption{
        \textbf{Training costs } --
        We report the cost of:
        1) Training the \apfg
        2) Training the RL agent
        3) Inference
        %
        %
        %
    }
    \label{tab:training-costs}
\end{table}

\subsection{Configuration Distribution}
\label{sec:eval-cfg-dist}

\begin{figure}
    \centering
    \begin{subfigure}[b]{0.6\columnwidth}
        \centering
        \includegraphics[width=0.8\linewidth]{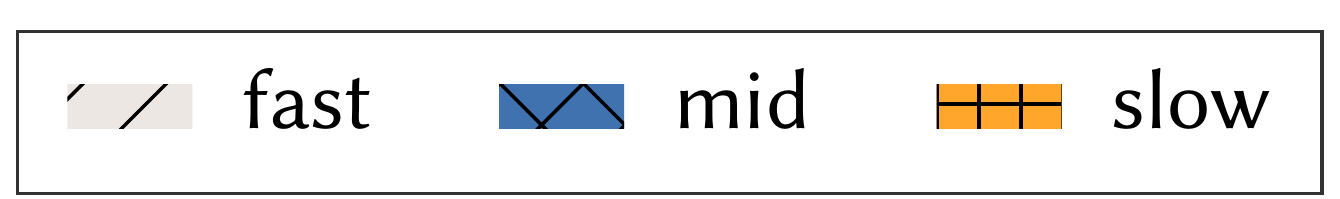}
        \includegraphics[width=\linewidth]{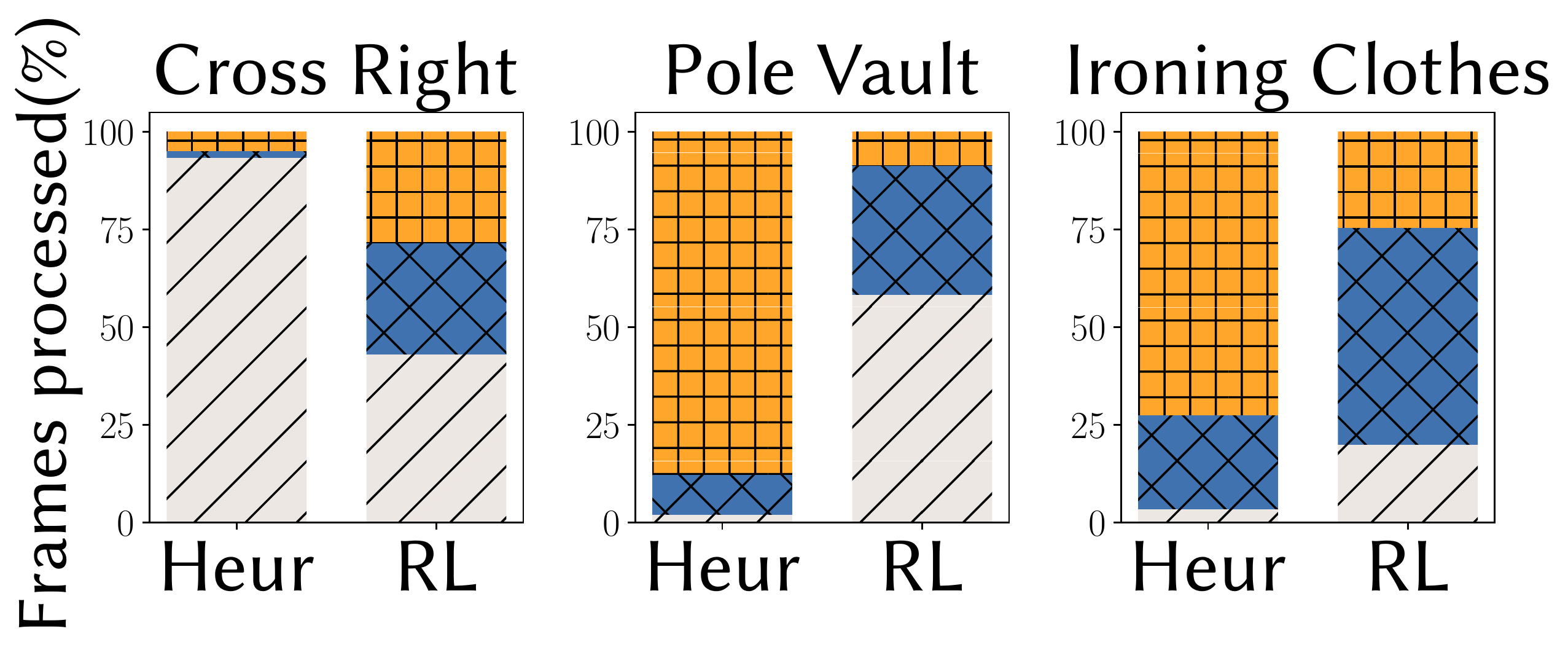}
        \caption{
         \textbf{Configuration distribution}
         }%
        \label{fig:config-dist}
    \end{subfigure}
    \hfill
    \begin{subfigure}[b]{0.35\columnwidth}
        \Huge
        \centering
        \begin{adjustbox}{width=\columnwidth,center}
        {\renewcommand{\arraystretch}{2}
        \begin{tabular}{ccc}
            \toprule
            \textbf{Query}  & \mbox{\textsc{Heuristic}\xspace}  & \mbox{\textsc{RL}\xspace}  \\ \midrule
            Cross Right  & 95/5 & 72/28\\ 
            Pole Vault   & 12/88 & 91/9\\
            Ironing Clothes & 28/72 & 75/25 \\ \bottomrule
        \end{tabular}
        }
    \end{adjustbox}
        \caption{
         Resolution (lo/hi(\%))
        }
        \label{tab:eval-resolutions}
    \end{subfigure}
       \caption{
           (a) Percentage of frames processed by the fast, mid, and slow configurations
                for \heuristic and \sysrl.
           (b) Percentage of frames processed at low and high resolutions in \heuristic and \sysrl.
           }
       \label{fig:eval-config-model}
\end{figure}

We next examine the efficacy of \sysrl in choosing the appropriate configurations
during query processing.
The configurations differ in both their throughput and accuracy of
processing queries (\cref{tab:example-configs}).
We constrain the agent to use three configurations:
(1) fast (but less accurate),
(2) mid, and
(3) slow (but more accurate).
The experimental settings are same as~\autoref{sec:eval-endtoend}.
We compare the percentage of frames processed using these
configurations for \heuristic and \sysrl.
The results are shown in~\cref{fig:config-dist}.

\sysrl processes the videos
using a combination of the configurations across all queries.
This differs from \heuristic that often uses a single configuration
for most of the frames.
On average, \heuristic processes 85\% of the frames using a single configuration.
In~\cref{fig:config-dist}, notice that \heuristic uses the fast configuration
for majority of frames in \crossright, while using the slow configuration for
majority of the frames in \polevault and \ironingclothes.
Since action percentage is low in \crossright, \heuristic uses the fastest
configuration for most of the frames, and in the process, skips important frames.
As a result, \heuristic falls well short (0.72) of the accuracy target (0.85).
Conversely, in \polevault and \ironingclothes, \heuristic reaches the target accuracy
but loses throughput (2$\times$ lower than \sysrl) due to the use of slow configurations.

In contrast, \sysrl uses a combination of the configurations in all the 3 queries.
It barely exceeds the target accuracy in both the queries and achieves
a high throughput.
%


\PP{Resolution Split}
We further divide the frames in~\cref{fig:config-dist} into low and high resolution frames,
based on the configuration used to process them.
\cref{tab:eval-resolutions} shows the percentage of frames processed at high and low
resolutions for \heuristic and \sysrl.
\heuristic uses a low resolution for majority of the frames in \crossright, since
it has a low action percentage.
On the other hand, it uses a high resolution in \polevault and \ironingclothes, which
have significantly higher action percentages.
We attribute this behavior to the rigidity of the rules used by \heuristic.
\sysrl on the other hand uses low resolution for majority of the frames in all three queries,
regardless of the action percentage in the videos.
This shows that the RL agent optimizes the throughput more efficiently.


%% file: related-work.tex
\section{Related Work}
\label{sec:related-work}


\PP{Action Localization}
Action Localization is a long-standing problem in computer vision.
Early efforts in AL focused on feature engineering
(\eg SIFT~\cite{lowe1999object}, HOG~\cite{dalal2005histograms}),
and hand crafted visual/temporal features~\cite{xia2020survey}.
To avoid hand-crafting features, deep neural networks have been proposed for AL.
Two-stream networks~\cite{simonyan2014two, feichtenhofer2016convolutional}
use deep neural networks to process inputs in rgb stream and optical flow stream.
3D Residual Convolutional Neural Networks (3D-CNNs)~\cite{tran2018closer}
use 3D-convolutions to directly process 4D-input blocks (stacked frames).
More recent work on AL includes SCNN~\cite{shou2016temporal}, and TAL-Net~\cite{chao2018rethinking}.
S-CNN and TAL-Net propose a three-stage approach for AL using proposal, classification,
and localization networks.
%
These above approaches focus on improving the \textit{accuracy} of AL.
On the other hand, \sys focuses on improving throughput of AL queries
while reaching a user-specified accuracy target.

\PP{Video Analytics}
Recent advances in vision have led to the development of numerous VDBMSs for
efficiently processing queries over videos.
\textsc{NoScope}~\cite{kang2017noscope} uses a cost-based optimizer to construct
a cascade of models (\eg lightweight neural network and difference detector).
Probabilistic Predicates (PP)~\cite{lu2018accelerating} are lightweight filters
that operate on frames to accelerate video classification.
These lightweight models accelerate query execution by filtering out frames
that do not satisfy the query predicate.
\textsc{BlazeIt}~\cite{kang13blazeit} extends filtering to more complex queries
(\eg aggregates and cardinality-limited
scrubbing queries). 
These lightweight filters accelerate execution by learning to directly answer
the query instead of using a heavyweight deep neural network.
 %
These lightweight filter-based methods cannot capture complex temporal and
scene information that is typically present in actions.
%
%
%
MIRIS~\cite{bastani2020miris} is a recently proposed VDBMS for processing 
object-track queries that requires processing a sequence of frames. 
It uses a graph neural network to keep track of objects between consecutive
frames.
%
%
It works well when a large number of frames satisfy the input query.
%

\sys differs from these efforts in that it is tailored to optimize action queries
while capturing the complex scene information.
It trains an \rl agent that adaptively chooses the input segments sent to the
3D-CNN.
Thus, it operates on a sequences of frames,  efficiently expresses complex scenes,
and handles rare events.


\PP{\rl for Video Processing}
Researchers have applied \rl for complex vision tasks such as video
summarization~\cite{zhou2018deep,lan2018ffnet,ramos2020straight}.
Deep Summarization Network (DSN)~\cite{zhou2018deep} uses deep \rl to generate
diverse and representative video summaries.
%
%
FastForwardNet (FFNet)~\cite{lan2018ffnet} uses deep \rl to
automatically fast forward videos and construct video summaries on the fly. 
%
%
%
%
In contrast, \sys uses \rl to process action queries.
Its optimizer leverages diverse knobs settings to reduce execution time.

%% file: conclusion.tex
\section{Conclusion}
\label{sec:conclusion}

Detecting and localizing actions in videos is an important problem in video
analytics. 
Current VDBMSs that use frame-based techniques are unable to answer action
queries since they do not extract context from a sequence of frames.
\sys processes action queries using a novel deep RL-based \proc. 
It automatically tunes three input knobs - resolution, segment length, and
sampling rate - to accelerate query processing by up to 4.7$\times$ compared to
state-of-the-art action localization techniques.
\sys uses an accuracy-aware \opt that generates aggregate rewards for training
the RL agent, ensuring that the \proc achieves the user-specified target
accuracy at a higher throughput than other baselines. 
%
%
%